\RequirePackage{fix-cm}

\documentclass[twocolumn]{svjour3}          % twocolumn
\smartqed  % flush right qed marks, e.g. at end of proof
\usepackage{natbib}
\usepackage{times}
\usepackage{epsfig}
\usepackage{graphicx}
\usepackage{subfigure}
\usepackage{amsmath,amssymb,amsfonts,dsfont,pifont,bm,bbm,mathrsfs,mathtools,nicefrac}
\usepackage{bm}
\usepackage{textcomp}
\usepackage{multirow}
\usepackage{xspace}
\usepackage{xcolor}
\usepackage{booktabs}
\usepackage{tabularx}
\usepackage{color}
\usepackage{makecell}
\usepackage{wrapfig}
\usepackage{bbding}
\usepackage{microtype}
\usepackage{ragged2e}
\justifying
\usepackage[ruled,linesnumbered]{algorithm2e}

\definecolor{citecolor}{RGB}{119,185,0} 
\usepackage[pagebackref=false,breaklinks=true,colorlinks,citecolor=citecolor,bookmarks=false]{hyperref}

\definecolor{upcolor}{RGB}{57,182,74}

\newlength\savewidth
\definecolor{Light}{RGB}{246,234,227}

\renewcommand{\paragraph}[1]{\vspace{1.25mm}\noindent\textbf{#1}}

\usepackage{amssymb}  % For \blacktriangle symbol
\usepackage{xcolor}   % For colored text

\definecolor{darkgreen}{RGB}{0,100,0}

\newcommand{\gain}[1]{%
  \textcolor{darkgreen}{\scriptsize $\blacktriangle$ \scriptsize #1}  %
}
\definecolor{darkred}{RGB}{100,0,0}

\newcommand{\drop}[1]{%
  \textcolor{darkred}{\scriptsize $\blacktriangledown$ \scriptsize #1}  
}

\makeatletter
\DeclareRobustCommand\onedot{\futurelet\@let@token\@onedot}
\def\@onedot{\ifx\@let@token.\else.\null\fi\xspace}

\usepackage{soul}
\soulregister\cite7
\soulregister\eqref7
\soulregister\ref7

\begin{document}

\title{GoClick: Lightweight Element Grounding Model for Autonomous GUI Interaction
}

\author{Hongxin~Li$^{1,2}$ \and
        Yuntao~Chen$^3*$ \and 
        Zhaoxiang~Zhang$^{1,2}*$
}

\institute{
    $^1$ University of Chinese Academy of Sciences, Beijing, China. \\
    $^2$ New Laboratory of Pattern Recognition, State Key Laboratory of Multimodal Artificial Intelligence Systems, Institute of Automation, Chinese Academy of Sciences, Beijing, China.
    \\
    $^3$ Hong Kong Institute of Science \& Innovation, Chinese Academy of Sciences, Hong Kong, China. \\  
    $^*$ Zhaoxiang Zhang and Yuntao Chen are the corresponding authors. \\
    E-mail: zhaoxiang.zhang@ia.ac.cn, chenyuntao08@gmail.com.
}

\date{Received: date / Accepted: date}

\maketitle

\begin{abstract}
Graphical User Interface (GUI) element grounding—precisely locating elements on screenshots based on natural language instructions—is fundamental for agents interacting with GUIs. Deploying this capability directly on resource-constrained devices like mobile phones is increasingly critical for GUI agents requiring low latency. However, this goal faces a significant challenge, as current visual grounding methods typically employ large vision-language model (VLM) ($\geq 2.5$B parameters), making them impractical for on-device execution due to memory and computational constraints. To address this, this paper introduces GoClick, a lightweight GUI element grounding VLM with only 230M parameters that achieves excellent visual grounding accuracy, even on par with significantly larger models. Simply downsizing existing decoder-only VLMs is a straightforward way to design a lightweight model, but our experiments reveal that this approach yields suboptimal results. Instead, we select an encoder-decoder architecture, which outperforms decoder-only alternatives at small parameter scales for GUI grounding tasks. Additionally, the limited capacity of small VLMs encourages us to develop a Progressive Data Refinement pipeline that utilizes task type filtering and data ratio adjustment to extract a high-quality 3.8M-sample core set from a 10.8M raw dataset. Training GoClick using this core set brings notable grounding accuracy gains. Our experiments show that GoClick excels on multiple GUI element grounding benchmarks while maintaining a small size and high inference speed. GoClick also enhances GUI agent performance when integrated into a device-cloud collaboration framework, where GoClick helps cloud-based task planners perform precise element localization and achieve higher success rates. We hope our method serves as a meaningful exploration within the GUI agent community.

\end{abstract}

\section{Introduction}
\label{sec:intro}
\justifying

\begin{figure}[t]
    \centering
    \includegraphics[width=1\linewidth]{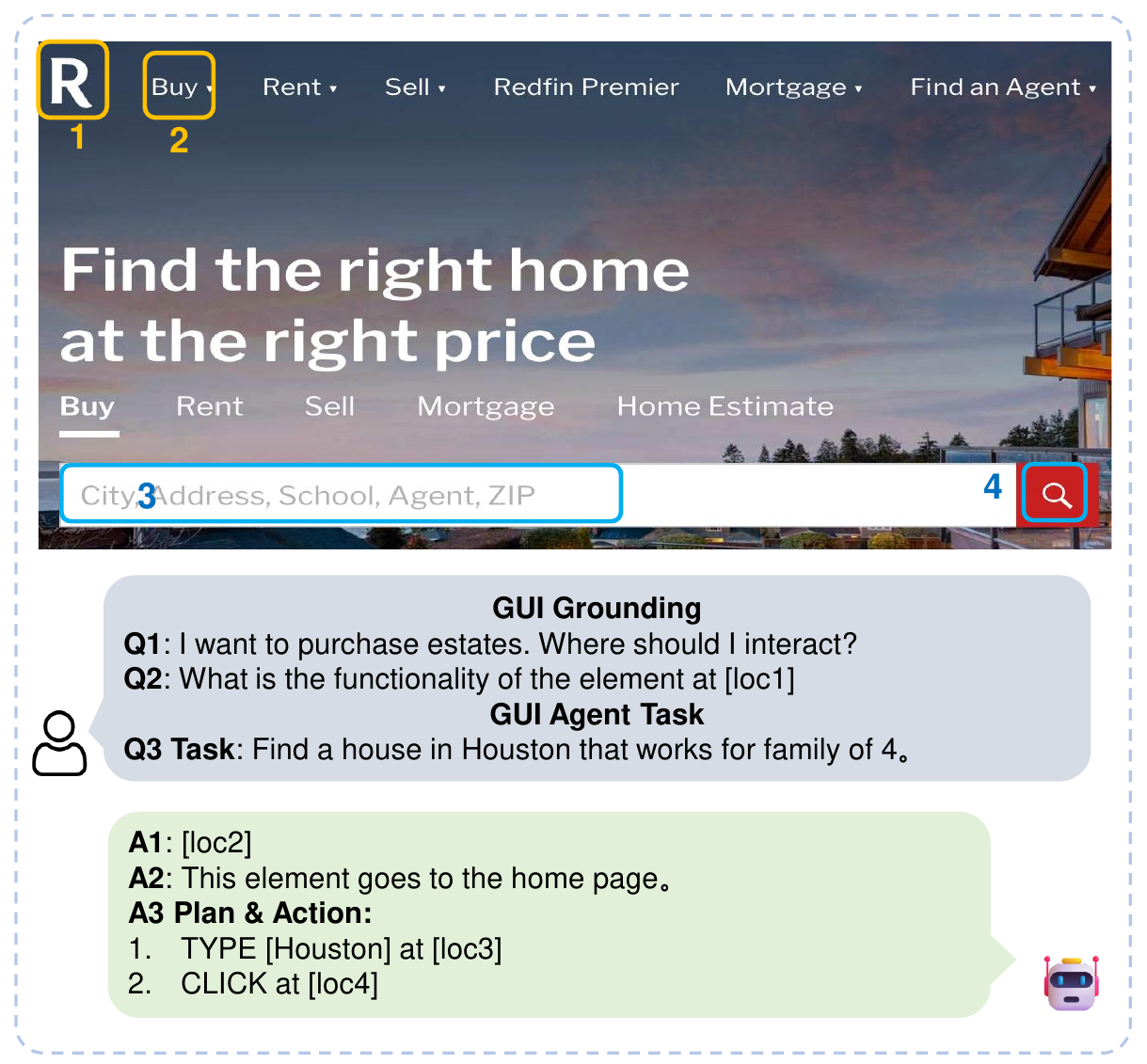}
    \caption{
    \textbf{Illustration of GUI interaction performed by a GUI agent capable of locating task-related elements.} This agent can assist users by automating operations on GUIs.}
    
    \label{fig:gui agent demo}
\end{figure}

% Resarch Topic Significance
% 
Users interact with digital applications through Graphical User Interfaces (GUIs) across mobile and web platforms.
Future GUI agents need to understand these interfaces to assist users in daily tasks (Fig.~\ref{fig:gui agent demo}).

A fundamental capability for such agents is \textit{GUI element grounding}—precisely locating elements on a GUI screenshot based on natural language instructions.
This capability enables GUI agents to accurately click elements according to instructions from visually impaired users and navigate mobile apps to collect information for human workers.
Deploying such capabilities on resource-constrained devices has become critical for applications requiring low latency.

% Technical Challenge
However, few GUI element grounding models have been tailored for these application scenarios.
Recent advancements in Vision-Language Models (VLMs)~\citep{qwen2vl,chen2024internvl} have driven the emergence of expert models with strong GUI element grounding capabilities~\citep{hong2023cogagent,cheng2024seeclick,uground}, but their deployment on resource-constrained devices remains challenging: these large models (often exceeding 2.5B parameters) are prohibitively expensive to run on mobile devices, creating tension between performance requirements and deployment constraints.

% Our solution
To address this challenge, we introduce \textbf{GoClick}, a light-weight GUI element grounding expert model with only 230M parameters that achieves excellent visual element grounding performance comparable to larger models while maintaining low latency.

% Method details 1: Model Arch.
A straightforward approach to developing a lightweight model is simply downsizing existing decoder-only VLMs, such as Qwen2VL~\citep{qwen2vl} and InternVL2~\citep{chen2024internvl}.
However, fine-tuning small-scale versions of these models (1B and 2B) yields only moderate grounding performance.
Instead, we select an encoder-decoder architecture, i.e. Florence-2~\citep{florence2}, as GoClick's base model and demonstrate that this architecture outperforms decoder-only alternatives at small parameter scales for GUI element grounding tasks.
\begin{figure}[ht]
    \centering
    \includegraphics[width=1\linewidth]{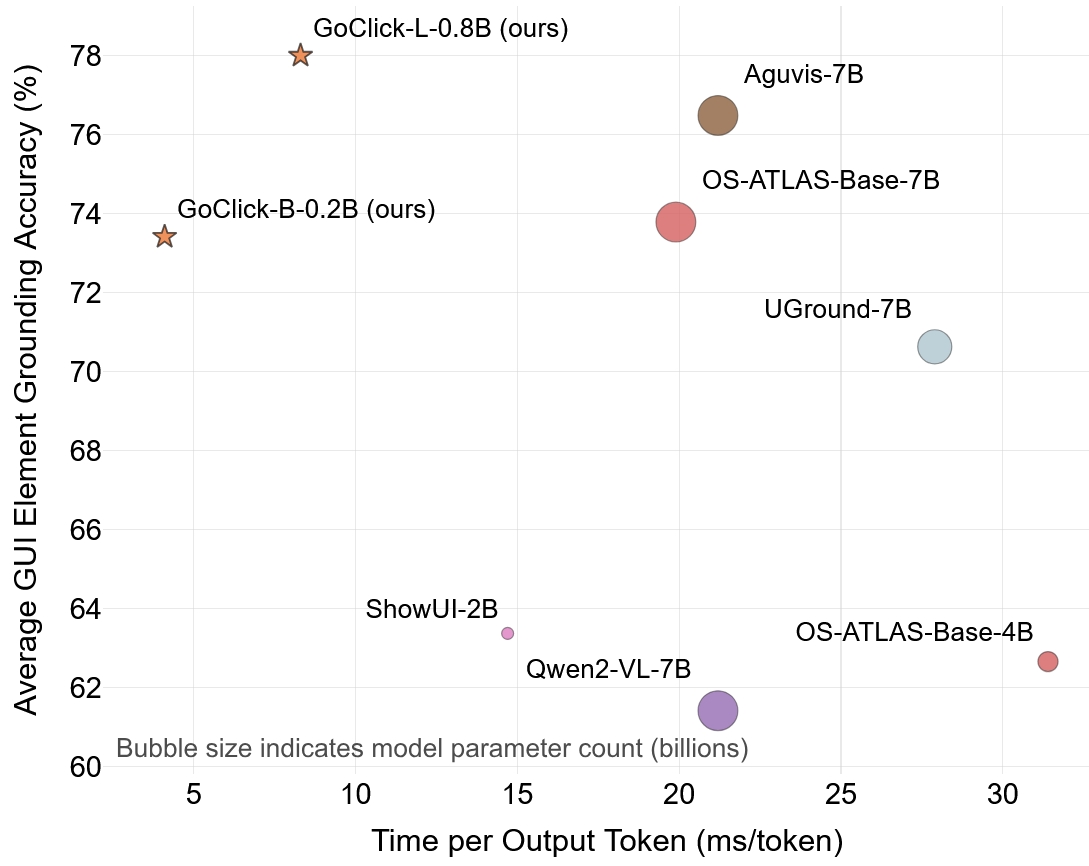}
    \caption{
    \textbf{Comparing GoClick with strong competitive VLMs in terms of inference speed, model size, and grounding accuracy on GUI element grounding benchmarks.} The average grounding accuracy shown on the Y-axis is calculated by averaging the accuracy on the seven benchmarks in Tab.~\ref{tab:gnd comparison}. Our GoClick achieves better average accuracy while enjoying a significantly smaller model size and higher inference speed.
    }
    \label{fig:speedacc tradeoff}
\end{figure}

% 改成星星

% Method details 2: Data refinement
Moreover, the limited capacity of small VLMs further motivates us to develop the Progressive Data Refinement (PDR) pipeline that extracts a high-quality core set from vast amounts of training data.
First, massive element referring expressions (REs) are annotated by rules and LLM-based annotators and then formatted as training samples with diverse element grounding task templates, resulting in a 10.8M raw dataset.
Next, We apply the PDR to retain useful samples:
in coarse-grained refinement, we exclude the samples with outdated GUI patterns and referring expression generation (REG) task samples as these samples harm GoClick's grounding performance;
in fine-grained refinement, we find that although the massive samples can bring performance gains, not all of them are helpful.
This finding drives us to reduce the inclusion ratios of tasks containing unhelpful samples to reach better grounding performance.
Our PDR ultimately yields a 3.8M high-quality core set used to train GoClick.

Extensive experiments show that on multiple GUI element grounding benchmarks, GoClick outperforms existing state-of-the-art models of equal size and performs comparably to several larger models (Fig.~\ref{fig:speedacc tradeoff}).

To demonstrate GoClick's practical applications, we evaluate it in a device-cloud collaboration agent framework, where cloud-based models (e.g., GPT-4o) handle high-level task planning while GoClick performs precise element localization on-device.
On GUI agent benchmarks focusing on operating digital devices, the device-cloud collaboration agent integrating GoClick achieves a substantially higher step success rate, outperforming both standalone proprietary models and Set-of-Marks~\citep{som} prompting strategies.
Our contributions are summarized as follows:

\begin{enumerate}
    \item We demonstrate that an encoder-decoder architecture outperforms decoder-only alternatives at small parameter scales for GUI element grounding tasks, suggesting a suitable lightweight base model for resource-constrained environments.
    \item We develop the Progressive Data Refinement pipeline that employs coarse-grained task type refinement and fine-grained task ratio adjustment to extract a high-quality training core set from massive raw data.
    \item By fine-tuning the selected encoder-decoder base model with this high-quality core set, we build GoClick as a lightweight GUI element grounding expert that surpasses existing equal-sized grounding models and compares favorably with several larger ones.
    \item GoClick exhibits promising applications in GUI agent tasks, where a device-cloud collaboration agent delegating grounding tasks to GoClick achieves higher step success rates compared to methods relying on proprietary models for grounding.
\end{enumerate}
\section{Related Works}
\label{sec:related works}

\subsection{Vision-Language Models}
There has been a significant rise in research focused on integrating both visual and textual inputs to perform multi-modal tasks~\citep{lxmert, alayrac2022flamingo, image_foreign,clip,liu2023llava,lin2023sphinx,chen2024internvl,lu2024deepseekvl,bai2023qwen,qwen2vl,zhu2024minigpt,li2023monkey,you2024ferret,peng2024kosmos,driess2023palme,cambrian}, which has led to the development of VLMs. Typically, VLMs are trained with visual instruction-tuning data to master a variety of complex tasks and to generate free-form responses leveraging visual information.

In the pre-GPT-4V era, several works~\citep{lxmert, vlbart, image_foreign, clip, alayrac2022flamingo} have proposed a spectrum of VLM architectures, which are primarily composed of the vision encoder and Transformer~\citep{Transformer}-based language models.
LXMERT~\citep{lxmert} and VL-BART~\citep{vlbart} explore leveraging a Transformer-based architecture to align and integrate visual features from images with linguistic features from text, enabling tasks like visual question answering and image captioning.
As a notable progress, Flamingo~\citep{alayrac2022flamingo} inserts trainable cross-attention layers between frozen Transformer layers of LLMs, thereby fusing language features with visual cues.
Trained with interleaved image-text data, Flamingo has demonstrated exceptional zero-shot task transfer and in-context learning capabilities.
Florence-2~\citep{florence2} is a Transformer model composed of encoder and decoder blocks that map visual-text input sequence to a textual output sequence. It achieves impressive multi-task capability spanning object detection, captioning, and grounding.
These VLM demonstrate better generalizability than traditional vision models designed for specific tasks.

With the development of LLMs, many works~\citep{alayrac2022flamingo, chen2023pali,Lin2023VILAOP,peng2024kosmos,driess2023palme,liu2023llava,lin2023sphinx,chen2024internvl,lu2024deepseekvl,bai2023qwen,qwen2vl,zhu2024minigpt,wang2024visionllm,li2023monkey,zhang2024llamaadapter,you2024ferret,cambrian,molmo} have explored infusing visual information into LLMs.
LLaVA~\citep{liu2023llava} is an end-to-end trained large vision-language model that connects a vision encoder (e.g., ViT~\citep{dosovitskiy2021vit}) with a pretrained LLM (e.g., Vicuna~\citep{vicuna}) via a visual-text projector, capable of general-purpose visual and language understanding.
Different from LLaVA, LLaMA-Adapter~\citep{zhang2024llamaadapter} does not use a projector but inserts learnable prompts into Transformer layers. These prompts are first embedded with visual inputs and then concatenated with text features for further feature fusion.
Models like VisionLLM~\citep{wang2024visionllm}, Ferret~\citep{you2024ferret}, and Qwen-VL~\citep{bai2023qwen} have further developed strong visual grounding abilities to locate target objects mentioned in referring expressions. Furthermore, LLaVA-Next~\citep{liu2024llavanext}, Monkey~\citep{li2023monkey}, InternVL-2,~\citep{InterVL-1.5}, Qwen2-VL~\citep{qwen2vl} are equipped with image division and capable of processing images of high resolution and any aspect ratio. Recently, research has been expanding to VLM applications in contexts with rich textual imagery~\citep{Tang2022UnifyingVT,2023-ureader,ye2023mplugdocowl,liu2024textmonkey} and embodied interactions~\citep{driess2023palme, mu2023embodiedgpt}, unlocking new possibilities in multimodal reasoning.
To provide practical guidelines, several works \citep{ mm15,idefics,molmo} also conduct comprehensive studies on building VLMs in terms of visual encoder design, data composition, and training recipes, along with open-sourced code base and data useful for the community.

These VLMs possess strong general multi-modal understanding capabilities, suggesting a potential path to building multi-modal autonomous agents. This work aims to improve GUI interaction task performance by training an element grounding model based on open-source VLMs.

\subsection{Multi-Modal GUI Agents}
Existing works~\citep{yao2022webshop,sheetcopilot,cheng2024seeclick,appagent,aguvis} have explored GUI agents capable of completing understanding and interaction tasks issued by human users.
Early studies focused on text-based GUI~\citep{yao2022webshop,synapse,sheetcopilot,deng2024mind2web}, investigating how to prompt proprietary LLMs (e.g., GPT-4) to operate GUIs to complete high-level tasks. Although these methods show promising outcomes of decision-making based on textual representations of GUIs, they are typically limited to narrow GUI scenarios that allow access to GUI source code.

A more practical way of building GUI agents is operating GUIs based on visual observations like humans~\citep{uground}.
This direction presents significant challenges due to the need for pixel-level control within a complex action space.
To address these challenges, recent studies have emphasized visually grounded GUI agents~\citep{hong2023cogagent,digirl, you2024ferretui, cheng2024seeclick, appagent, uground, osatlas,aguvis,omniparser}.
A straightforward way is building upon proprietary VLMs, such as GPT-4V~\citep{openai2024gpt4}.
MM-Navigator~\citep{yan2023gpt}, AppAgent~\citep{appagent}, SeeAct~\citep{seeact}, Mobile-Agent~\citep{mobileagent} are developed to unleash the GUI interaction capability of GPT-4V by either crafting role-playing prompts or adopting Set-of-Marks prompting~\citep{som}.
Although these methods can utilize the general reasoning ability of GPT-4V, it is hard to further improve the agent performance by simple prompt engineering.

Recent works have also explored training GUI agents based on open-source VLMs.
Ferret-UI~\citep{you2024ferretui}, SeeClick~\citep{cheng2024seeclick}, CogAgent~\citep{hong2023cogagent}, OS-ATLAS~\citep{osatlas}, and Aguvis~\citep{aguvis} have generated massive multi-modal instruction-following tasks to train open-source VLMs as GUI agents.
Additionally, some works~\citep{omniparser, uground} also combine proprietary VLMs (e.g. GPT-4V~\citep{openai2024gpt4} and PaLM2~\citep{anil2023palm2technicalreport}) as task decomposers while training open-source VLMs as low-level action executors.
These methods employ the reasoning ability of proprietary VLMs and the specialized execution ability of trained VLMs to obtain better performance.

However, these models are typically large VLMs ($\geq$ 7B) that cannot be easily deployed on mobile devices.
In contrast, this work is dedicated to developing a lightweight yet strong GUI element grounding model tailored to GUI agents in resource-constrained scenarios.
A recent work, OmniParser-v2~\citep{omniparser}\footnote{The details of OmniParser-v2 can be found at https://github.com/microsoft/OmniParser}, also trains lightweight VLMs to assist in GUI interaction, but it focuses on GUI element detection and captioning, which is different from the GUI element grounding emphasized in this work.
\section{Methodology}
\label{sec:method}

\subsection{Preliminary}

\subsubsection{GUI Element Grounding Task}
GUI element grounding is to predict either the box coordinates $c = [x_{topleft},y_{topleft},x_{bottomright},y_{bottomright}]$ or the center coordinates $c = [x_{center},y_{center}]$ of the target GUI element $e$ within a GUI screenshot image $I$ given a natural language referring expression (RE) $L$:
\begin{equation}
    c = g(I, L; \varphi)
\end{equation}
where $\varphi$ denotes the parameters of the grounding model $g$. This task enables natural language interaction with GUIs by bridging user intentions and interface elements.

Common REs include: (a) Spatial references: ``the button at the top right‘’, (b) Appearance-based references: ``the blue submit button'', (c) Functional references: ``the login field'', and (d) Combined references: ``the red close button in the settings menu''. Refer to Fig.~\ref{fig:gnd example}.

Unlike natural scene visual grounding~\citep{refcoco}, GUI element grounding introduces unique challenges:
1) High resolution: GUIs require models that process large visual inputs efficiently.
2) Fine-grained comprehension: GUIs contain numerous small elements occupying less area than objects in datasets like RefCOCO~\citep{refcoco} and Visual Genome (VG)~\citep{visualgenome}, demanding enhanced visual understanding to distinguish similar elements.

\begin{figure*}[t]
    \centering
    \includegraphics[width=1\linewidth]{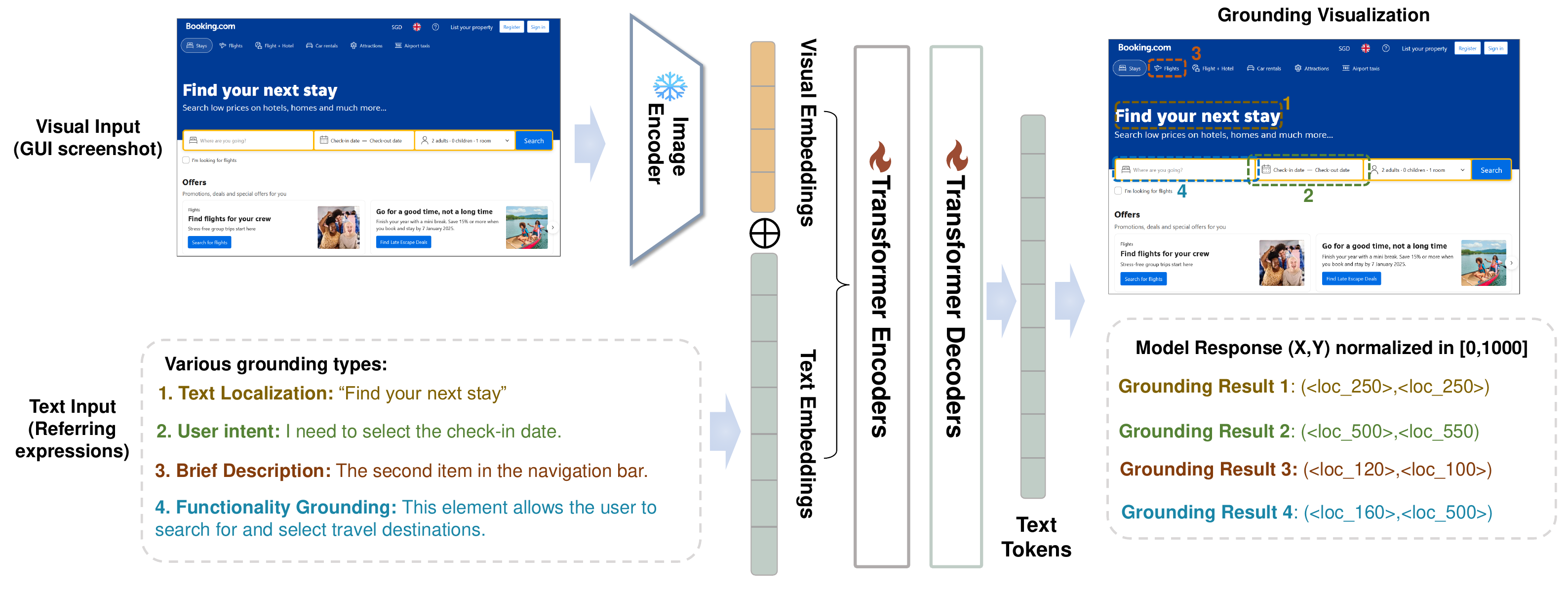}
    \caption{
    \textbf{\textit{GoClick}} is based on Florence-2~\citep{florence2}, which constitutes an image encoder and a multi-modality encoder-decoder module. We fine-tune Florence-2 with the collected GUI element grounding tasks, leading to a small yet strong expert model in locating target elements according to various referring expressions, such as texts, action intents, functionality descriptions, and brief visual descriptions.
    }
    \label{fig:model_arch}
\end{figure*}

\subsubsection{GUI Agent Task}
GUI agent tasks involve complex, multi-step interactions following high-level user instructions.
Example tasks include: (a) Navigation Tasks: ``Open the settings menu and enable dark mode'' and ``Go to my profile and update the contact information''; (b) Form Completion: ``Fill out the registration form with my information'' and ``Book a flight to New York for next Friday''; (c) Information Retrieval: ``Find the price of the cheapest item in the shopping cart'' and ``Check if there are any unread messages''. Refer to Fig.~\ref{fig:gui agent task example} for examples.

Since GUI navigation requires element interaction, robust element grounding capability is crucial for GUI agent task success~\citep{omniparser,uground}.
The grounding model must accurately localize elements given various RE types, mapping natural language descriptions to precise screen coordinates.
This task is challenging due to language nuances (appearance descriptions, contextual dependencies) and visual similarity among elements in complex layouts.
On mobile devices, these difficulties are compounded by computational constraints requiring lightweight yet accurate models capable of real-time inference.
Addressing these challenges would improve the operation smoothness of GUI agents.

In this paper, we develop a lightweight yet strong GUI element grounding model based on a small VLM and deploy it in GUI agent tasks.

\subsection{GoClick Model Architecture}
We adopt Florence2's encoder-decoder architecture~\citep{florence2} as the base model of our \textbf{GoClick}, shown in Fig.~\ref{fig:model_arch}.
This architecture combines a pre-trained visual encoder $f_{\phi}$  (e.g., ViT~\citep{dosovitskiy2021vit}) and Transformer~\citep{Transformer}-based language model. the visual encoder maps an input image $\displaystyle \mathbf{I} \in \mathbb{R}^{H\times W \times 3}$ to an $L$-length sequence of patch features $\mathbf{V}_{img} \in \mathbb{R}^{L\times h}$ which are projected into visual token embeddings.
Subsequently, the visual embeddings $E_{img} \in \mathbb{R}^{L\times D}$ are concatenated with $S$-length textual embeddings $E_{txt} \in \mathbb{R}^{S\times D}$ before being fed into the multi-modal encoder. The decoding process can be formulated as
\begin{equation}
    o = \text{Encoder-Decoder}_{\theta}([Proj_{\omega}(f_{\phi}(\displaystyle \mathbf{I})), Embed(\displaystyle \mathbf{t}) ]
\end{equation}

where $\displaystyle \mathbf{t}$, $Proj$, and $Embed$ denote the text prompt, the vision-language projector, and the LLM embedding module, respectively.
For each training sample ($\displaystyle \mathbf{I}$, $\displaystyle \mathbf{t}$, $o$), the VLM is optimized by minimizing the loss $L(\phi, \omega, \theta) = -\log p(o | \displaystyle \mathbf{I}, \displaystyle \mathbf{t})$ through gradient descent.

We adopt this encoder-decoder architecture instead of a decoder-only one like Qwen2-VL~\citep{qwen2vl} as prior research~\citep{ul2,pmlr-v162-wang22u} demonstrates the performance advantages of encoder-decoder architectures in a narrow task scenario at a small parameter scale.
This choice aligns with our focus on visual element grounding, which requires precise spatial localization rather than extensive free-form generation. The encoder specializes in visual-linguistic feature extraction, while the compact decoder focuses solely on coordinate prediction—a narrower output space than open-ended text generation. This design avoids the overhead of maintaining broad conversational capabilities, which are generally required for decoder-only models. The advantage of our design choice will be justified in Sec.~\ref{sec:experiments}.

\subsection{Data Compilation Engine}

\begin{figure*}[ht]
    \centering
    \includegraphics[width=1\linewidth]{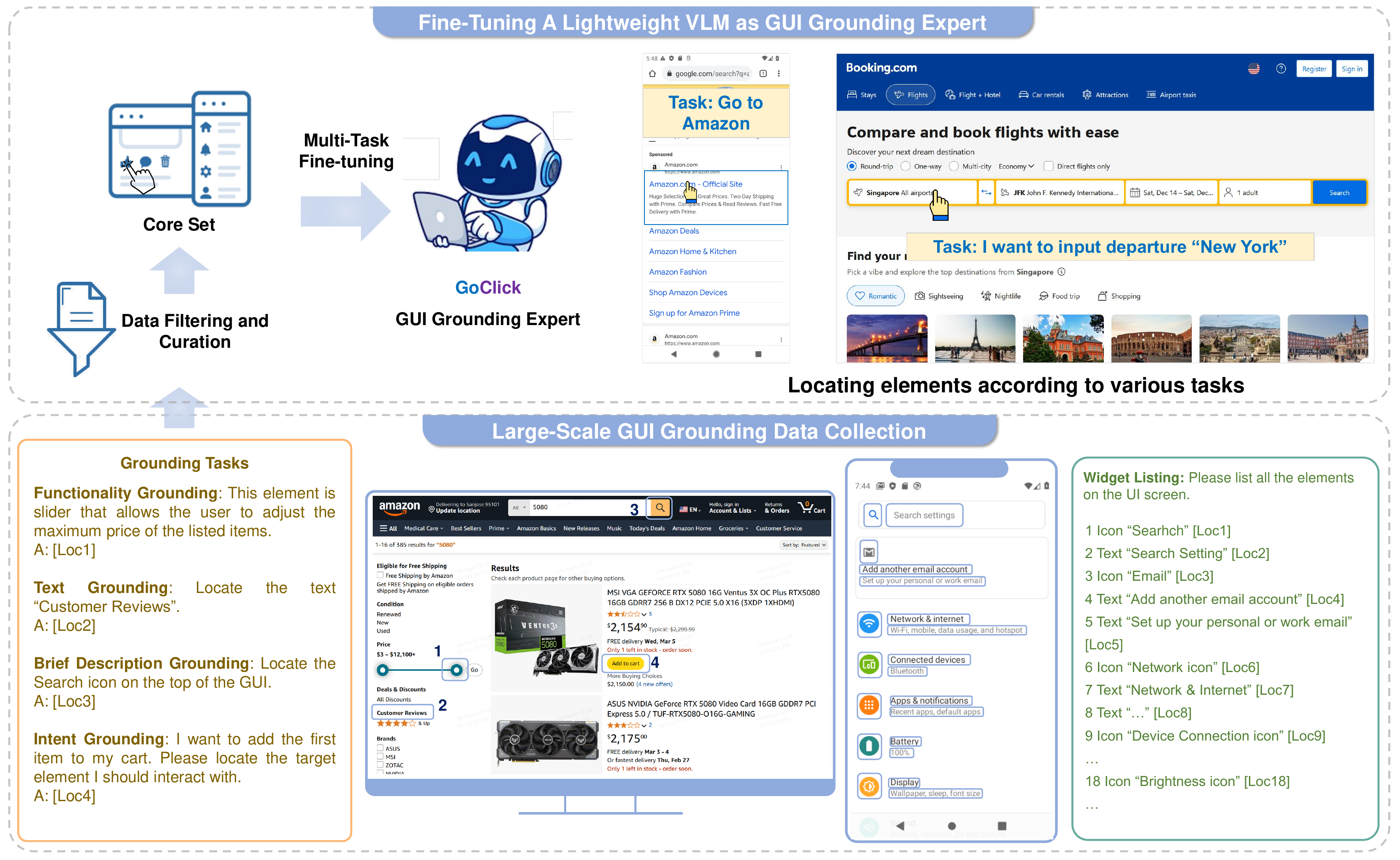}
    \caption{
    \textbf{Illustration of the data collection and multi-task fine-tuning processes of GoClick.} To efficiently foster a strong GUI grounding, we first curate a large-scale GUI grounding training data containing diverse task types and then conduct procedural filtering and curation to extract a high-quality core set. Finally, this core set is used to train GoClick as a lightweight GUI grounding expert capable of performing accurate element grounding given various referring expressions, such as action intents.
    }
    \label{fig:data engine}
\end{figure*}

Training GoClick as a GUI grounding expert requires a comprehensive, high-quality dataset spanning diverse GUI domains and element REs.
While previous works~\citep{hong2023cogagent,cheng2024seeclick,osatlas} acknowledge the benefits of massive data collection, the limited capacity of GoClick drives us to develop the Progressive Data Refinement (PDR) pipeline to select high-quality portions.
This subsection first introduces raw data collection and then details the proposed PDR pipeline.

\subsubsection{Collecting REs for GUI Elements}
\label{sec: RE type intro}
Unlike abundant language datasets (e.g., BookCorpus~\citep{bookcorpus} and RefinedWeb~\citep{refineweb}) and natural image collections (e.g., ImageNet~\citep{Russakovsky2014ImageNetLS} and LAION5B \citep{LAION5B}), GUI element RE data remains scarce, primarily due to the complex annotation requirements for high-resolution screenshots and GUI expertise demand.

To address this insufficiency, we curate a large-scale dataset through GUI metadata annotation and diverse grounding task generation.
Each grounding task is represented by a \textless screenshot, RE, coordinates\textgreater triplet. We generate various four types of REs:

\noindent{\textbf{RE1: Alt-texts}} - Basic element descriptors, including displayed text for plain-text elements  (such as paragraphs and headings) and descriptive text for images/icons (such as a magnifier icon and a figure in an article).

\noindent{\textbf{RE2: Brief descriptions}} - Enhanced descriptions incorporating visual appearance, element category, and positional information. For example, a house-shape element might be described as ``a navigating-home button at the top left''.

\noindent{\textbf{RE3: Action intent}} - User interaction descriptions following~\cite{Burns2022ADF}, such as ``focus on the Password textbox'' for a password input field.

\noindent{\textbf{RE4: Functionality descriptions}} - Descriptions of the interactive affordance of GUI elements~\citep{autogui}. For example, an element shaped like an upward arrow might be annotated with ``This element enables users to share content with others''. This RE type helps models comprehend the functional semantics of elements, complementing the aforementioned three types of REs that are derived from the element attributes in GUI source code.

We directly generate the former three REs from GUI metadata, and follow AutoGUI pipeline~\citep{autogui} to generate abundant functionality descriptions using LLM annotators.%
Note that multiple REs can be generated for the same element. After obtaining the triplets, we generate massive GUI grounding tasks, including \textbf{Text Grounding}, \textbf{Brief Description Grounding}, \textbf{Intent Grounding}, and \textbf{Functionality Grounding} based on the alt-texts, brief descriptions, action intents and functionality annotations, respectively.

We also implement reverse tasks (REG - Referring Expression Generation) for all types except Intent Grounding, following previous work~\citep{cheng2024seeclick,hong2023cogagent}. These REG tasks are Text REG, Brief Description REG, and Functionality REG that prompt the model to generate REs given target element locations.
Additionally, to further enhance grounding capability, we follow Ferret-UI~\citep{you2024ferretui} to generate \textit{widget listing} samples, which requires detecting all visible elements.

Our dataset draws from diverse GUI sources (Tab.~\ref{tab:coreset task portion}), including multi-resolution webpages from AutoGUI~\citep{autogui} and WebUI~\citep{WebUI}, mobile GUIs from various emulated Android devices~\citep{autogui,mobileagent}, and training splits from GUI agent task benchmarks~\citep{androidcontrol,rawles2023android,guicourse,deng2024mind2web}.

\subsubsection{Denoising And De-Duplication Procedure}
\label{sec: denoise}
GUI data may contain plenty of noise and duplication due to GUI design defects. Training with the GUI grounding tasks generated based on these unignorable noisy elements will likely hamper model performance and waste computational resources.

To resolve this issue, we check the used GUI data sources, recognize element noise types, and develop a denoising pipeline to clean the raw data: a) Discard blank and invisible elements; b) Remove elements with invalid bounding boxes that exceed image borders; c) Use Optical Character Recognition (OCR) tools~\footnote{https://pypi.org/project/pytesseract} to remove plain-text elements whose displayed texts do not match the text attributes due to GUI rendering faults. This denoising procedure helps to ensure the final GoClick fine-tuning dataset is generated based on clean elements.

Apart from noise, redundancy is also a critical problem. Samples with highly similar element REs and positions are likely to appear in abundance as GUI designs are probably similar across apps. For example, navigating-back icons at the top-left and search bars at the top appear frequently. These redundant similar samples lead to diminishing training returns while increasing training time.

To address this issue, a simple yet effective approach is adopted: we first normalize and discretize the grounding boxes of all elements in a coarse integer range $[0-100]$, then clean the REs of these elements by removing all punctuation marks and lowering all letters, and finally group all elements by the discretized boxes and cleaned REs (i.e. Elements with identical discretized box coordinates and REs fall in the same group). After grouping, we randomly select only one sample from each group and discard the remaining samples. This approach results in a reduction of redundant samples, significantly decreasing the dataset size while maintaining diversity.

We ultimately collect a raw dataset of \textbf{10.8M} training samples (GoClick-Raw).

\subsubsection{Progressive Data Refinement Pipeline}
\label{sec:core set curation}

Despite denoising and de-duplication, our dataset remains large and imbalanced across task types from various sources.
Here, we further develop the progressive data refinement (PDR) pipeline to extract a high-quality core set. To reduce the complexity of investigating all task types at all ratios, PDR adopts a coarse-to-fine refining procedure:

\noindent\textbf{Coarse Refining Stage} We first evaluate two categories of potentially unhelpful samples:

\begin{enumerate}
    \item Non-element grounding tasks (e.g., REG tasks that generate descriptions from element locations)
    \item Samples featuring outdated GUI design patterns that differ significantly from contemporary mobile devices and benchmarks. For instance, WAE~\citep{chen2020wireframe} contains GUI screenshots from Android 4.0 (2011), while our experimental benchmarks (Sec.~\ref{sec:gnd bmks}) use Android versions post-2017.
\end{enumerate}
 
The two sample categories are empirically shown to be unhelpful in the experiments (Sec.~\ref{sec:exp data ablation}) and are removed from the fine-tuning data.
This initial refinement yields a 6.8M-sample subset (GoClick-AfterCoarse).

\noindent\textbf{Fine Refining Stage} We then dive into the fine-grained adjustment of the combination of the remaining tasks.
Inspired by previous works on data balancing~\citep{cambrian, xu2024demystifying}, we systematically reduce the volume of each task type while monitoring performance on grounding evaluation benchmarks
If one task type is essential, then reducing its amount likely results in a notable performance drop.

To isolate the effects of image diversity and task type ratio, we conduct adjustments for each GUI metadata source. However, it is time-consuming to investigate the task ratios for all sources. Hence, we focus on the six major sources that constitute 90.8\% of the 6.8M-sample dataset, while retaining all samples from minor sources.

After removing the useless data portions that contribute negatively to grounding performance, we finally obtain a 3.8M-sample high-quality core set.

\subsubsection{Discussion}
While data filtering has been investigated~\citep{ xu2024demystifying} in the VLM field, our PDR pipeline contributes these novel aspects specifically tailored to GUI visual grounding tasks:
First, systematic investigation of GUI-specific harmful patterns: We identify that outdated GUI patterns (e.g., Android 4.0 from 2011) harm performance on contemporary benchmarks. This finding is non-obvious, as one might assume that older data simply provide additional training signal.

Second, task-type level analysis that reveals counterintuitive patterns: Our fine-grained refinement (Fig.~\ref{fig:fine filtering}) reveals that not all grounding tasks contribute equally, and some popular sources actually hurt performance when fully included. For example, SeeClick-Web shows a catastrophic performance decline at full inclusion.

Third, quantifiable efficiency gains: Our PDR reduces training data from 10.8M to 3.8M samples (reduction of 64.8\%) while improving the average precision (Tab.~\ref{tab:data ablation overview}). This represents a fundamental contribution to data efficiency: achieving superior performance with substantially fewer samples and reduced training time.

\subsubsection{Dataset Preview And Statistics}

% \usepackage{booktabs}
% \usepackage{graphicx}
\begin{table*}[ht]
\centering
\caption{\textbf{The number of samples for every task type in the GoClick core set generated by the proposed Progressive Data Refinement pipeline in Sec.~\ref{sec:core set curation}.} The asterisk (*) denotes the main GUI metadata sources that occupy 90.8\% of the 6.8M-size GoClick-AfterCoarse set and undergo the task ratio ablation studies for the fine filtering stage in Sec.~\ref{sec:exp data ablation}.}
\label{tab:coreset task portion}
\resizebox{\textwidth}{!}{%
\begin{tabular}{@{}cc|cccccc@{}}
\toprule
Metadata Source & Domain & Text Gnd. & Brief Desc. Gnd. & Intent Gnd. & Functionality Gnd. & Widget Listing & \textbf{Total} \\ \midrule
AutoGUI*~\citep{autogui} & Web, Mobile & 18k & 101k & 33k & 376k & 77k & 604k \\
MobileViews*~\citep{mobileviews} & Mobile & 39k & - & 104k & - & 52k & 195k \\
AndroidControl*~\citep{androidcontrol} & Mobile & - & - & - & 23k & 9k & 32k \\
WebUI*~\citep{WebUI} & Web & - & 41k & 79k & - & 34k & 154k \\
MultiUI*~\citep{yu2015multi} & Web, Mobile & - & 646k & 1221k & - & - & 1868k \\
SeeClickWeb*~\citep{cheng2024seeclick} & Web & - & 617k & - & - & - & 617k \\
GUIEnv~\citep{guicourse} & Web & 50k & - & - & - & - & 50k \\
RICO~\citep{deka2017rico} & Mobile & - & 42k & 225k & - & - & 266k \\
Mind2Web~\citep{deng2024mind2web} & Web & - & - & 14k & - & - & 14k \\
AITW~\citep{rawles2023android} & Mobile & - & - & 11k & - & - & 11k \\
OmniAct~\citep{kapoor2024omniact} & Web, Desktop & - & 3k & - & - & - & 3k \\
\textbf{Total} & - & \textbf{107k} & \textbf{1450k} & \textbf{1686k} & \textbf{399k} & \textbf{172k} & \textbf{3814k} \\ \bottomrule
\end{tabular}%
}
\end{table*}
% \usepackage{booktabs}
% \usepackage{graphicx}
\begin{table*}[t]
\centering
\caption{\textbf{The statistics of the 3.8M core set}. \#Total Tokens calculates the total number of the tokens of the 3.8M prompt-response pairs. The number of unique elements is greater than the number of samples as one widget listing sample involves multiple elements.}
\label{tab:data stats}
\resizebox{\textwidth}{!}{%
\begin{tabular}{@{}cccccccc@{}}
\toprule
Item &
  \begin{tabular}[c]{@{}c@{}}\#GUI Metadata\\ Sources\end{tabular} &
  \#Samples &
  \#Unique Elements &
  \begin{tabular}[c]{@{}c@{}}\#Unique GUI\\ Screenshots\end{tabular} &
  \#Total Tokens &
  \begin{tabular}[c]{@{}c@{}}\#Average Prompt\\ Tokens\end{tabular} &
  \begin{tabular}[c]{@{}c@{}}Top-4 Common\\ Resolutions\end{tabular} \\ \midrule
Value &
  11 &
  3,814,466 &
  4,833,823 &
  967,975 &
  218,440,537 &
  29.1 &
  \begin{tabular}[c]{@{}c@{}}``$1920 \times 1080$'': 618k, ``$1080\times 1920$'': 401k\\  ``$2560\times 1440$'': 297k, ``$1284\times 2238$'': 201k\end{tabular} \\ \bottomrule
\end{tabular}%
}
\end{table*}

The task ratios and statistics of the 3.8M core set are illustrated in Tab.~\ref{tab:coreset task portion} and Tab.~\ref{tab:data stats}. Examples of the grounding tasks are shown in Fig.~\ref{fig:data engine}.

\section{Experiments}
\label{sec:experiments}
\begin{figure*}[ht]
    \centering
    \includegraphics[width=1\linewidth]{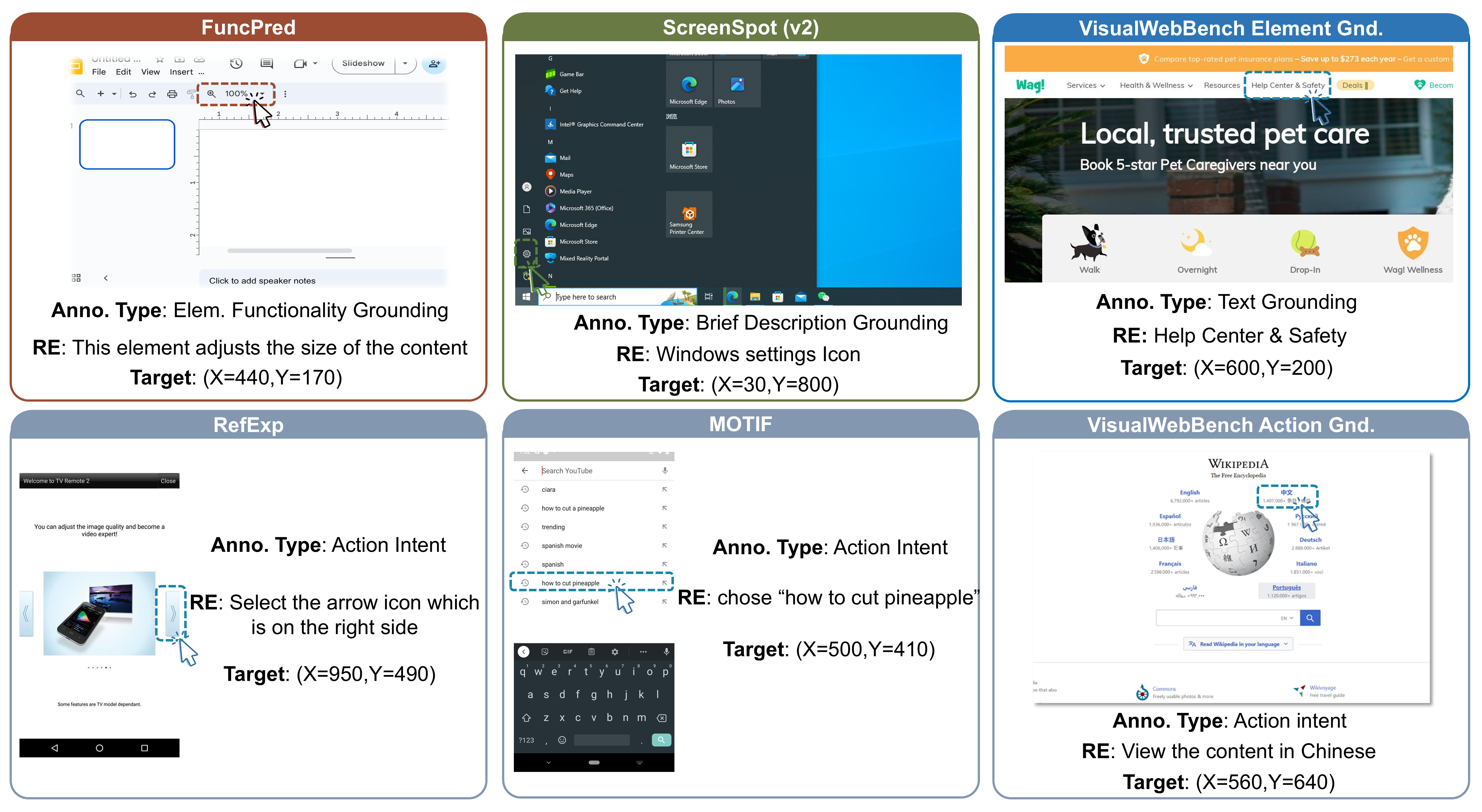}
    \caption{Examples of the grounding tasks provided by the used GUI element grounding benchmarks in Sec.\ref{sec:gnd bmks}.}
    \label{fig:gnd example}
\end{figure*}

This section presents our experiments that (1) demonstrate the better speed-accuracy tradeoff of GoClick (Sec.~\ref{sec:comp sota}), (2) compare GoClick with variants built upon decoder-only VLMs to justify our choice of architecture (Sec.~\ref{sec:exp model arch}), (3) investigate the benefits brought by the proposed core set curation pipeline (Sec.~\ref{sec:exp data ablation}), and (4) show how our grounding expert model assists in GUI agent tasks (Sec.~\ref{sec:exp 2stage planning}).

\subsection{Comparisons with State-of-The-Art Methods}
\label{sec:comp sota}
To demonstrate the better speed-accuracy tradeoff obtained by our GoClick, we compare it with existing leading VLMs capable of GUI element grounding.

\subsubsection{The Used GUI Element Grounding Benchmarks}
\label{sec:gnd bmks}
For a solid assessment of the models, we conduct a comprehensive evaluation on multiple GUI element grounding benchmarks featuring diverse GUI domains and various types of REs:

\begin{enumerate}
    \item \textbf{FuncPred}~\citep{autogui} is the test split of the AutoGUI dataset~\citep{autogui}. This benchmark requires a model to locate the element specified by its contextual functionality description. This benchmark is hard as the REs in this benchmark focus on the functionality of the target elements and do not mention visual or positional characteristics.
    \item \textbf{ScreenSpot}~\citep{cheng2024seeclick} and \textbf{ScreenSpot-v2}~\citep{osatlas} comprise test samples on mobile, desktop, and web platforms. They require the model to locate elements based on brief descriptions of the elements.
    \item \textbf{MOTIF}~\citep{Burns2022ADF} requires a model to locate elements given a natural language intent on mobile apps. Each intent in this dataset mentions the appearance, category, and position of a target element (e.g., \textit{click the icon at the top right corner}).
    \item \textbf{RefExp}~\citep{Bai2021UIBertLG} also requires the model to locate elements given action intents. Each intent also explicitly mentions the appearance, category, and position of a target element.
    \item \textbf{VisualWebBench}~\citep{liu2024visualwebbench} is a comprehensive multimodal benchmark assessing the understanding capabilities of VLMs in web scenarios. We select the two grounding tasks from this benchmark, i.e. element grounding (VWB EG) and action grounding (VWB AG).
\end{enumerate}

Examples of these benchmarks are displayed in Fig.~\ref{fig:gnd example}.
For all of these benchmarks, we report \textbf{Grounding Accuracy (\%)}, which denotes the percentage of samples with the predicted points falling within the bounding boxes of the target elements:
\begin{equation}
    \text { Acc }=\sum_{i=1}^N \mathbf{1}\left(\text {pred}_i \text { inside GT } \text { bbox }_i\right) / N \times 100
\end{equation}
where $\mathbf{1}$ is an indicator function and $N$ the number of test samples.

We also assess the inference speeds using two popular metrics~\citep{ondevice_survey}:

\noindent \textbf{a) Time to First Token (TTFT)}. This metric shows how long a user has to wait before seeing the model’s output. This is the time it takes from submitting the query to receiving the first token (if the response is not empty). 

\noindent \textbf{b) Time per Output Token (TPOT)}. This is defined as the average time between consecutive tokens\footnote{The definition of TPOT follows the manual in https://docs.nvidia.com/nim/benchmarking/llm/latest/metrics.html}:
\begin{equation}
    \text{TPOP} = \frac{\text{e2e\_latency} - \text{TTFT}}{\text{Total\_output\_tokens} - 1}
\end{equation}

where $\text{e2e\_latency}$ indicates how long it takes from submitting a query to receiving the full response. This does not include the first token (hence subtracting 1 in the denominator) in order to consider only the decoding part of the request processing.

\subsubsection{Experimental Settings}
We compare with general-purpose VLMs, including GPT-4o and Qwen2VL~\citep{qwen2vl}, and previously leading VLMs trained with GUI element grounding tasks, including SeeClick~\citep{cheng2024seeclick}, Ferret-UI~\citep{you2024ferretui}, UGround~\citep{uground}, OS-ATLAS~\citep{osatlas}, and Aguvis~\citep{aguvis}.
Most of these models are built on large VLMs with more than 7B parameters.

As these models cannot be easily deployed on mobile devices, the inference speed metrics are calculated under conditions simulating mobile device usage.
Each model processes an input image (height: 780, width: 360) and a prompt of median length derived from the grounding benchmark test samples (Sec.~\ref{sec:gnd bmks}). Then, the model processes the image and the prompt 2000 times to calculate the average of TTFT and TPOT. One L20 GPU is used for evaluation, with a batch size being only 1. Model inference is conducted using the Transformers library~\citep{wolf-etal-2020-transformers} with Flash Attention~\citep{dao2022flashattention, dao2023flashattention2} where supported. All models use BFloat16 precision and deterministic generation (temperature = 0.0) without sampling.

\subsubsection{Experimental Results}
% % Please add the following required packages to your document preamble:
% % \usepackage{booktabs}
% \usepackage{graphicx}
\begin{table*}[]
\centering
\caption{\textbf{Comparing GoClick with existing GUI element grounding models on the grounding benchmarks.} GoClick-L outperforms the models smaller than 7B. Compared with the 7B competitive models, GoCliCK-L surpasses these models on FuncPred, MOTIF, RefExp, and VWB EG while achieving comparable accuracy on ScreenSpot and VWB AG. GoClick-B, with merely 0.2B parameters, still obtains remarkably high grounding accuracy. F, B, I, and T (The referring expression type used in the benchmarks) denote Functionality Grounding, Brief Description Grounding, Intent Grounding, and Text Grounding, respectively. M, W, and D (GUI domains) denote mobile, web, and desktop scenarios.}
\label{tab:gnd comparison}
\resizebox{\textwidth}{!}{%
\begin{tabular}{@{}cccc|ccccccc@{}}
\toprule
Model &
  Size &
  \begin{tabular}[c]{@{}c@{}}TTFT $\downarrow$\\ (ms)\end{tabular} &
  \begin{tabular}[c]{@{}c@{}}TPOT $\downarrow$\\ (ms/token)\end{tabular} &
  \begin{tabular}[c]{@{}c@{}}FuncPred\\ (F; M, W)\end{tabular} &
  \begin{tabular}[c]{@{}c@{}}ScreenSpot\\ (B; M, W, D)\end{tabular} &
  \begin{tabular}[c]{@{}c@{}}ScreenSpot-v2\\ (B; M, W, D)\end{tabular} &
  \begin{tabular}[c]{@{}c@{}}MOTIF\\ (I; M)\end{tabular} &
  \begin{tabular}[c]{@{}c@{}}RefExp\\ (I; M)\end{tabular} &
  \begin{tabular}[c]{@{}c@{}}VWB EG\\ (T; W)\end{tabular} &
  \begin{tabular}[c]{@{}c@{}}VWB AG\\ (I; W)\end{tabular} \\ \midrule
GPT-4o                                                   & -    & - & - & 9.8       & 17.8      & 20.4      & 30.5      & 21.8      & 5.6       & 6.8       \\
Qwen2VL-7B~\citep{qwen2vl}            & 8B   & 118.9 & 21.2 & 38.7      & 66.4      & 66.9      & 75.1      & 64.8      & 55.9      & 62.1      \\
CogAgent~\citep{hong2023cogagent}  & 18B  & 1253.2  & 208.8  & 29.3      & 47.4      & 49.2      & 46.7      & 35.0      & 55.7      & 59.2      \\
SeeClick~\citep{cheng2024seeclick} & 10B  & 160.4  & 184.4 & 19.8      & 53.4      & 54.0      & 11.1      & 58.1      & 39.2      & 27.2      \\
Ferret-UI~\citep{you2024ferretui}   & 8B   & 152.5  & 22.9  & 1.2       & 7.1       & 7.8       & 15.9      & 5.5       & 3.9       & 1.9       \\
UGround~\citep{uground}     & 7B   & 1034.6  &  27.9 & 48.8      & 74.8      & 76.5      & 72.4      & 73.6      & 85.2      & 63.1      \\
OS-ATLAS-8B~\citep{osatlas}       & 8B   & 137.5 & 19.9 & 52.1      & 82.5      & 84.1      & 78.8      & 66.5      & 82.6      & 69.9      \\
Aguvis~\citep{aguvis}    & 8B   & 119.7  & 21.2  & 52.0       & 83.8      & 85.6      & 73.8      & 80.9      & 91.3     & 68.0     \\
\midrule
Qwen2-VL~\citep{qwen2vl}    & 2B   &  58.8 & 16.4  & 7.1       & 17.9      & 18.6      & 28.8      & 29.2      & 17.9      & 17.5      \\
OS-ATLAS-4B~\citep{osatlas}       & 4B   & 137.3  &  31.4 &  44.6  &  66.8   & 68.7    &   75.4   &  77.1   &    47.7  &  58.3   \\
Ferret-UI~\citep{you2024ferretui}   & 3B   & 69.5  & 9.8  & 1.3       & 2.1       & 1.9       & 5.5       & 1.1       & 0.7       & 1.0       \\
ShowUI~\citep{showui}    & 2B   & 79.7  & 14.7 & 39.9       & 76.1     & 77.4      & 72.3      & 58.4      &  64.2      & 55.3      \\

GoClick-L (ours)                                               & 0.8B & 91.1  &  8.3 & \textbf{69.5} & \textbf{78.5} & \textbf{81.1} & \textbf{80.4} & \textbf{78.2} & \textbf{90.3} & \textbf{68.0} \\
GoClick-B (ours)                                             & 0.2B &  \textbf{37.7} & \textbf{4.1}  &   64.4   &     74.1  &  75.2  &  76.8  &  71.9   &   90.3   &   61.2  \\ \bottomrule
\end{tabular}%
}
\end{table*}

The results in Tab.~\ref{tab:gnd comparison} and Fig.~\ref{fig:speedacc tradeoff} show that GoClick achieves a significantly better tradeoff between inference speed and grounding accuracy. GoClick not only surpasses the competitive models with model sizes less than 4B but also achieves performance comparable to stronger expert models with 7B parameters.
Notably, GoClick-B-0.2B (230M), with approximately 1/30 parameters, 1/3 TTFT, and 1/5 TPOT, even outperforms several competitive 7B models on FuncPred, VWB EG, and obtains comparable accuracy on MOTIF and RefExp.
Moreover, our GoClick is fine-tuned with \textbf{only 3.8M} samples, substantially fewer than the 13.6M used by OS-ATLAS~\citep{osatlas} and 10M by UGround~\citep{uground}.

These results indicate that the synergy of the adopted encoder-decoder architecture and the proposed core set curation pipeline enhances the parameter efficiency and data efficiency of GoClick.

\subsection{Experiment on Model Architecture}
\label{sec:exp model arch}
To evaluate the effectiveness of our encoder-decoder architecture for GUI grounding, GoClick is compared to the variants that are based on state-of-the-art decoder-only VLM architectures, focusing on GUI element grounding accuracy.

\subsubsection{Experimental Settings}

We compare with variants based on the decoder-only VLMs in this experiment:

\textbf{InternVL-2}~\citep{chen2024internvl} adopts a LLaVA-like architecture (``ViT-MLP-LLM''), with Qwen-2.5 as its decoder-only LLM. Similarly, InternVL-2.5-1B splits high-resolution images into $448 \times 448$ patches and encodes them separately, then employs a pixel unshuffle operation to reduce the number of visual tokens to one-quarter of the original. This VLM is also further trained with massive multi-image and video instruction-following data.

\textbf{Qwen2-VL}~\citep{qwen2vl} also adopts ``ViT-MLP-LLM'' architecture, with Qwen-2 as its decoder-only LLM. Similarly, Qwen2-VL-2B can also process high-resolution GUI screenshot images by splitting an image into patches and then encoding them separately before inputting them into the LLM. This VLM is also further trained with massive multi-image and video instruction-following data.

\textbf{SLIME}~\citep{slime}: This VLM equips the LLaVA model~\citep{liu2023llava} with an image division mechanism that supports efficient processing of high-resolution images. It also uses a local patch resampling method to reduce visual tokens, thereby improving inference speed while reducing noise.

We use the small-scale versions of these models for comparison, i.e. InternVL-2-1B, Qwen2-VL-2B, and SLiME-Gemma-2B. Our model and the three compared decoder-only VLMs are all fine-tuned with our 3.8M-size core set for two epochs. For training efficiency and fair comparison, all GUI screenshot images are resized to $768 \times 768$, as we empirically find that the key details (e.g., texts and icon patterns) of GUI screenshots are still discernible at this resolution. For all models, only the ViT encoder is frozen. All fine-tuning experiments are conducted on eight L20 GPUs. Fine-tuning GoClick-L-0.8B, Qwen2-VL-2B, InternVL-2-1B, and SLiME-Gemma-2B frozen takes 38, 24, 9, and 18 hours, respectively.

While these models were pretrained on different datasets, comparing them after fine-tuning with our GUI element grounding dataset remains valid and informative. This is because: (1) These models have been pretrained with massive grounding tasks (e.g., RefCOCO~\citep{coco} and FLD~\citep{florence2}), so they possess a basic capability of outputting numeric coordinates. This ensures that none of these models has to learn to output numeric coordinates from scratch; (2) By fine-tuning all models under identical conditions (same epochs and fine-tuning data), we can evaluate how effectively each architecture adapts to the unique challenges of GUI element grounding.
These two points help to ensure that final performance differences primarily reflect each architecture's inherent ability to learn precise spatial grounding rather than pretraining advantages.

\subsubsection{Experimental Results}

% \usepackage{booktabs}
% \usepackage{graphicx}
% \usepackage{booktabs}
% \usepackage{multirow}
% \usepackage{graphicx}
% \usepackage{multirow}
% \usepackage{graphicx}
\begin{table*}[]
\centering
\caption{\textbf{Comparing grounding accuracy (\%) of GoClick based on the adopted encoder-decoder architecture and decoder-only alternatives.} The adopted encoder-decoder architecture, i.e. Florence2-Large~\citep{florence2}, and the compared decoder-only models are fine-tuned with our 3.8M core set for two epochs and then evaluated on the various GUI element grounding benchmarks. GoClick based on Florence2 remarkably outperforms the variant fine-tuned from InternVL-2-1B with an equal model size. Although Qwen2VL-2B and SLiME-Gemma-2B possess two times the parameters of GoClick, they also underperform GoClick after fine-tuning. Notably, GoClick achieves comparable performances on ScreenSpot-v2, MOTIF, RefExp, and VWB, compared with the fine-tuned Qwen2-VL-7B, and outperforms it on FuncPred. These results suggest the advantage of adopting the Florence2 encoder-decoder architecture as the base model of GoClick for GUI element grounding tasks.}
\label{tab:arch comp}
\resizebox{\textwidth}{!}{%
\begin{tabular}{c|ccccccccc}
\hline
 &
  Model &
  Size &
  \begin{tabular}[c]{@{}c@{}}FuncPred\\ (F; M, W)\end{tabular} &
  \begin{tabular}[c]{@{}c@{}}ScreenSpot\\ (B; M, W, D)\end{tabular} &
  \begin{tabular}[c]{@{}c@{}}ScreenSpot-v2\\ (B; M, W, D)\end{tabular} &
  \begin{tabular}[c]{@{}c@{}}MOTIF\\ (I; M)\end{tabular} &
  \begin{tabular}[c]{@{}c@{}}RefExp\\ (I; M)\end{tabular} &
  \begin{tabular}[c]{@{}c@{}}VWB EG\\ (T; W)\end{tabular} &
  \begin{tabular}[c]{@{}c@{}}VWB AG\\ (I; W)\end{tabular} \\ \hline
\multirow{5}{*}{Base}  & InternVL2-8B       & 8.1B & 2.7 & 5.3 & 4.9 & 15.2 & 3.7 & 1.9 & 4.9 \\ 
                    & Qwen2VL-7B       & 8.3B & 38.7 & 66.4 & 66.9 & 75.1 & 64.8 & 55.9 & 62.1 \\ \cline{2-10}
                     & SLiME-Gemma-2B   & 2.8B & 4.2  & 13.0 & 13.4 & 7.0  & 8.3  & 6.1  & 4.9  \\
                     & Qwen2VL-2B       & 2.2B & 7.1  & 17.9 & 18.6 & 28.8 & 29.2 & 17.9 & 17.5 \\
                      & InternVL-2-1B    & 0.9B & 2.0  & 3.4  & 3.0  & 6.3 & 3.5  & 0.7  & 2.9  \\
                     & Florence2-L        & 0.8B & 2.3  & 4.3  & 4.6  & 10.4 & 9.0  & 1.5  & 5.8  \\ \hline
\multirow{5}{*}{SFT} 
                     & InternVL2-8B     & 8.1B & 28.8 & 51.4 & 51.0 & 70.4 & 77.7 & 29.1 & 26.2 \\
                     & Qwen2VL-7B      & 8.3B & 56.1 & 82.5 & 86.7 & 79.1 & 81.4 & 93.5 & 67.0 \\ \cline{2-10} 
                     
                     & SLiME-Gemma-2B   & 2.8B & 58.9 & 56.8 & 58.4 & 69.8 & 50.3 & 39.5 & 33.0 \\
                     & Qwen2VL-2B       & 2.2B & 51.1 & 73.5 & 76.5 & 56.3 & 74.8 & 78.2 & 62.1 \\
                      & InternVL-2-1B    & 0.9B & 21.9 & 21.9 & 22.2 & 50.8 & 68.3 & 7.3 & 7.8 \\
                     & GoClick-L (ours) & 0.8B & \textbf{69.5} & \textbf{78.5} & \textbf{81.1} & \textbf{80.4} & \textbf{78.2} & \textbf{90.3} & \textbf{68.0} \\ \hline
\end{tabular}%
}
\end{table*}
\begin{table}[]
\centering
\caption{\textbf{Evaluating GoClick on the subsets of the ScreenSpot benchmark}~\citep{cheng2024seeclick}. GoClick based on Florence2's encoder-decoder architecture achieves a robust grounding performance across the three GUI domains (i.e., Web, Desktop, and Mobile) and also enjoys a smaller gap between the easier text grounding and the harder icon grounding performances compared with the variant trained from the strong decoder-only Qwen2-VL-2B. Note that the overall performance in the last column corresponds to the ScreenSpot performance in Tab.~\ref{tab:arch comp}.}
\label{tab:sspot detail arch abate}
\resizebox{\columnwidth}{!}{%
\begin{tabular}{@{}cc|cc|cc|cc|c@{}}
\toprule
\multirow{2}{*}{Base Model} & \multirow{2}{*}{Model Size} & \multicolumn{2}{c|}{Mobile} & \multicolumn{2}{c|}{Desktop} & \multicolumn{2}{c|}{Web} & \multirow{2}{*}{Overall} \\
                     &      & Text & Icon & Text          & Icon & Text         & Icon &      \\ \midrule
SLiME-Gemma-2B       & 2.8B & 77.3 & 43.2 & 68.0          & 32.1 & 63.0         & 41.8 & 56.8 \\
Qwen2-VL-2B          & 2.2B & 85.4 & 48.5 & 87.1          & 56.4 & 83.0         & 61.2 & 71.5 \\
InternVL-2-1B        & 0.9B & 43.6 & 32.3 & 20.1 & 10.7 & 8.7 & 5.8  & 21.9 \\
Florence2-L (GoClick-L) & 0.8B & \textbf{88.6} & \textbf{70.3} & \textbf{84.5}          & \textbf{67.1} & \textbf{86.5}         & \textbf{67.5} & \textbf{78.5} \\ \bottomrule
\end{tabular}%
}
\end{table}

The results in Tab.~\ref{tab:arch comp} demonstrate the superiority of our GoClick model built upon the encoder-decoder architecture of Florence-2~\citep{florence2}, compared to competitive decoder-only VLM alternatives.

\noindent \textbf{a) GoClick achieves overall performance superiority.}
With approximately the same number of parameters, our GoClick (0.8B) based on the encoder-decoder Florence-2 outperforms InternVL-2.5-1B-SFT on all the benchmarks. GoClick also surpasses Qwen2VL-2B-SFT and SLiME-Gemma-2B-SFT across all benchmarks.
Particularly notable is GoClick’s performance on FuncPred, where it achieves an accuracy of 69.5\%, significantly surpassing the compared models.

\noindent \textbf{b) Comparison with Larger Decoder-Only Models.} 
Surprisingly, despite having only one-tenth of the parameters, GoClick (0.8B) achieves comparable or superior results compared to Qwen2VL-7B-SFT (8.3B).
For example, GoClick outperforms Qwen2VL-7B-SFT on FuncPred (+13.4) and VWB AG (+1.9) and achieves similar performance on Screen-Spot-v2, MOTIF, and RefExp.
This highlights the parameter-efficiency advantage of fine-tuning GoClick based on the encoder-decoder Florence2 architecture.

\noindent \textbf{c) Better cross-domain robustness.}
GoClick exhibits more robust performance across diverse benchmarks that span various RE types and GUI domains.
For instance, GoClick achieves impressive grounding accuracy on Brief Description Grounding (e.g., ScreenSpot), Intent Grounding (e.g., MOTIF and RefExp), and even the difficult Functionality Grounding (i.e., FuncPred). 
Moreover, the ScreenSpot performance split into three different GUI domains in Tab.~\ref{tab:sspot detail arch abate} shows that GoClick enjoys a stronger and more robust grounding capability across domains.

Overall, the results indicate the advantage of building GoClick upon Florence-2's encoder-decoder architecture.

% \usepackage{booktabs}
% \usepackage{graphicx}
\begin{table*}[]
\centering
\caption{
\textbf{The effects of the coarse filtering stage in Sec.~\ref{sec:core set curation}.}
We remove the samples with outdated GUI patterns and samples of each referring expression generation (REG) task to investigate whether these samples are useful. The results show that removing outdated samples leads to slightly higher grounding accuracy. Removing the REG task samples either brings performance gains or remains approximately unchanged performance, which indicates that adding the dual REG tasks which share the same target elements of the original grounding tasks does not benefit GoClick's grounding capability. Func. REG, Text REG, and Brief Desc. REG denote functionality description generation, text generation, and brief description generation grounded at target elements, which are introduced in Sec.~\ref{sec: RE type intro}.
}
\label{tab:corase filtering}
\resizebox{\textwidth}{!}{%
\begin{tabular}{@{}cccccccc@{}}
\toprule
Data Abaltion &
  \begin{tabular}[c]{@{}c@{}}FuncPred\\ (F; M, W)\end{tabular} &
  \begin{tabular}[c]{@{}c@{}}ScreenSpot\\ (B; M, W, D)\end{tabular} &
  \begin{tabular}[c]{@{}c@{}}ScreenSpot-v2\\ (B; M, W, D)\end{tabular} &
  \begin{tabular}[c]{@{}c@{}}MOTIF\\ (I; M)\end{tabular} &
  \begin{tabular}[c]{@{}c@{}}RefExp\\ (I; M)\end{tabular} &
  \begin{tabular}[c]{@{}c@{}}VWB EG\\ (T; W)\end{tabular} &
  \begin{tabular}[c]{@{}c@{}}VWB AG\\ (I; W)\end{tabular} \\ \midrule
GoClick-Raw         & 62.2 & 72.2 & 74.4 & 74.0 & 72.0 & 86.4 & 60.2 \\ \midrule

w/o Old GUI Patterns & 62.4 \gain{0.2} & 72.2 \gain{0.0} & 74.6 \gain{0.2} & 76.8 \gain{2.8} & 74.9 \gain{2.9} & 88.4 \gain{2.0} & 62.1 \gain{1.9} \\ 
w/o Func. REG        & 63.1 \gain{0.9} & 74.4 \gain{2.2} & 76.9 \gain{2.5} & 75.1 \gain{1.1} & 74.2 \gain{2.2} & 90.2 \gain{3.8} & 61.2 \gain{1.0} \\
w/o Text REG     & 62.2 \gain{0.0} & 73.4 \gain{1.2}  & 75.3 \gain{0.9} & 74.9 \gain{0.9} & 72.2 \gain{0.2} & 87.2 \gain{0.8} & 61.2 \gain{1.0} \\
w/o Brief Desc. REG  & 62.5 \gain{0.3} & 73.1 \gain{0.9} & 74.0 \drop{0.4} & 75.8 \gain{1.8} & 72.3 \gain{0.3} & 88.9 \gain{2.5} & 60.2 \gain{0.0} 
%w/o Elem. Class.     & 62.3 \gain{0.1} & 72.3 \gain{0.1} & 74.2 \drop{0.2} & 74.2 \gain{0.2} & 71.7 \drop{0.3} & 86.4  \gain{0.0} & 61.2  \gain{1.0}
\\
% \midrule

% After Coarse Filtering     & 63.0 \gain{0.8} & 74.9 \gain{2.7} & 76.3 \gain{1.9} & 77.2 \gain{3.2} & 75.6 \gain{3.6} & 90.1  \gain{3.7} & 61.2  \gain{1.0} \\

\bottomrule
\end{tabular}%
}
\end{table*}

\subsection{Experiment on Fine-Tuning Data Refinement}
\label{sec:exp data ablation}
This subsection presents the experimental settings, results, and analysis for the proposed PDR pipeline in Sec.~\ref{sec:core set curation}.

\subsubsection{Experimental Settings}

We follow the procedure in Sec.~\ref{sec:core set curation} to conduct coarse refinement and then fine refinement to curate a much smaller yet more effective core set of the original GoClick full set.
GoClick-L is used for this experiment. To address computational constraints, each fine-tuning experiment is run for one epoch across the extensive experimental matrix.
Given that our primary objective is to perform a comparative analysis of relative performance trends between different data ratios and task types, a single-epoch training regimen provides a sufficient signal to identify meaningful performance differentials while saving computational budget.

\subsubsection{Experimental Results}

\begin{figure*}[t]
    \centering
    \includegraphics[width=1\linewidth]{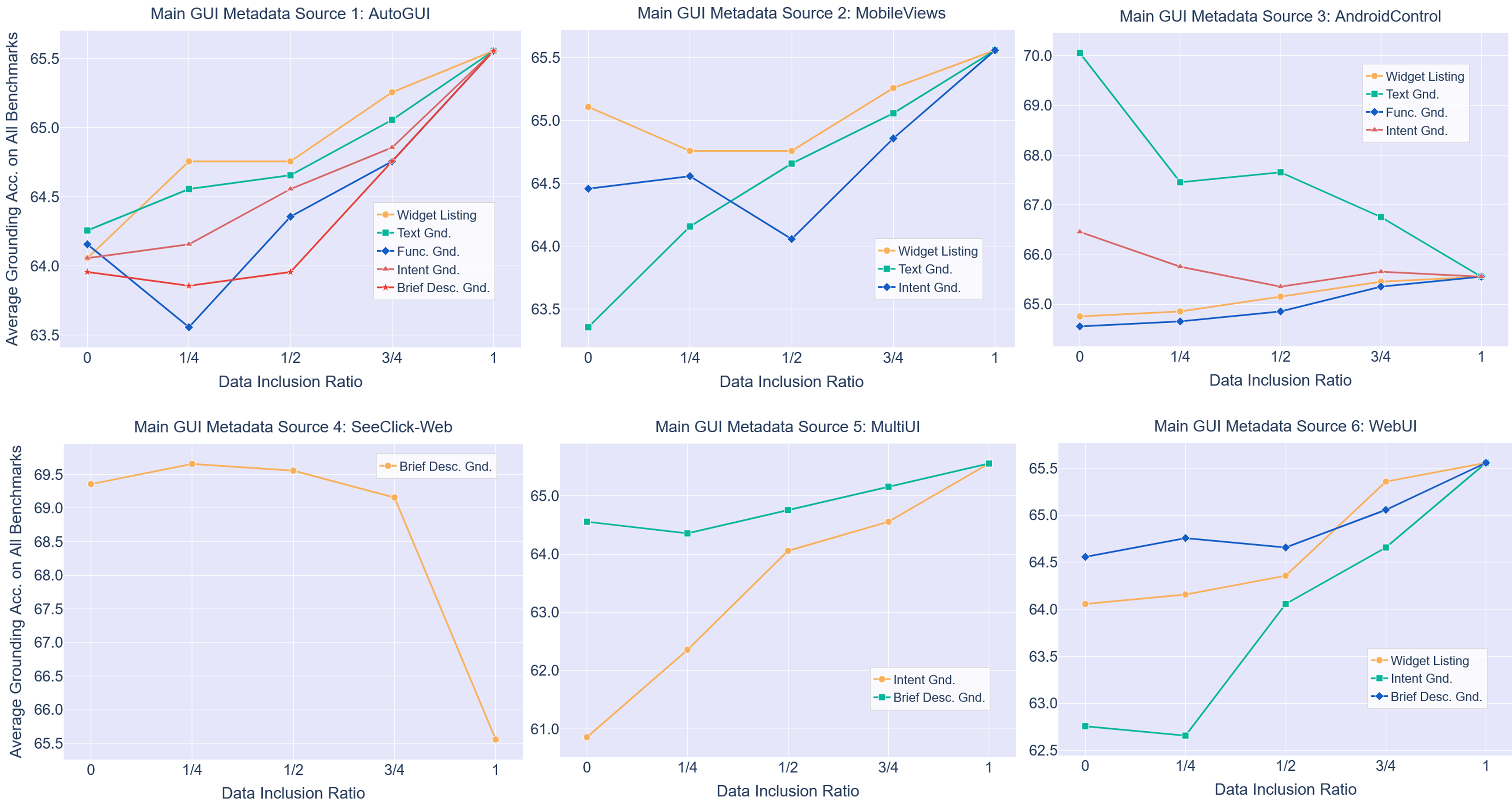}
    \caption{
     \textbf{The effects of adjusting the inclusion ratio of each task type for the main GUI metadata sources in the fine filtering stage described in Sec.~\ref{sec:core set curation}}. We investigate the efficacy of each task type by gradually decreasing the inclusion ratio to observe grounding performance variations. For all of the subplots, the data point at ratio=1 denotes the performance when all of the main sources (except the minor sources) are used to fine-tune the base model, Florence2. The curves show that not all data are useful for improving GUI element grounding accuracy. The tasks generated from four of the six sources (i.e., AutoGUI, MobileViews, MUltiUI, and WebUI) are useful. However, the Text Grounding and Intent Grounding tasks in AndroidControl, as well as the Brief Description Grounding tasks in SeeClick-Web do not positively contribute to grounding performance when all are included. These results suggest that we can reduce the inclusion ratios of the unhelpful task samples to achieve higher grounding performance.
    }
    \label{fig:fine filtering}
\end{figure*}
% \usepackage{booktabs}
% \usepackage{graphicx}
\begin{table*}[]
\centering
\caption{\textbf{The effects of the two filtering stages introduced in Sec.~\ref{sec:core set curation}.} Coarse filtering leads to notable performance gains while significantly reducing training samples. Fine filtering brings a further increase. These results demonstrate the efficacy of the proposed core set curation pipeline.}
\label{tab:data ablation overview}
\resizebox{\textwidth}{!}{%
\begin{tabular}{@{}cccccccccc@{}}
\toprule
Model &
  \#Samples &
  \begin{tabular}[c]{@{}c@{}}FuncPred\\ (F; M, W)\end{tabular} &
  \begin{tabular}[c]{@{}c@{}}ScreenSpot\\ (B; M, W, D)\end{tabular} &
  \begin{tabular}[c]{@{}c@{}}ScreenSpot-v2\\ (B; M, W, D)\end{tabular} &
  \begin{tabular}[c]{@{}c@{}}MOTIF\\ (I; M)\end{tabular} &
  \begin{tabular}[c]{@{}c@{}}RefExp\\ (I; M)\end{tabular} &
  \begin{tabular}[c]{@{}c@{}}VWB EG\\ (T; W)\end{tabular} &
  \begin{tabular}[c]{@{}c@{}}VWB AG\\ (I; W)\end{tabular} &
  Average \\ \midrule
GoClick-Raw &
  10.8M &
  62.2 &
  72.2 &
  74.4 &
  74.0 &
  72.0 &
  86.4 &
  60.2 &
  71.6 \\
After Coarse Filtering &
  6.4M &
  63.0 \gain{0.8} &
  74.9 \gain{2.7} &
  76.3 \gain{1.9} &
  77.2 \gain{3.2} &
  75.6 \gain{3.6} &
  90.1 \gain{3.7} &
  61.2 \gain{1.0} &
  74.0 \gain{2.4} \\
After Fine Filtering &
  3.8M &
  64.5 \gain{2.3} &
  75.3 \gain{3.1} &
  77.5 \gain{3.1} &
  78.3 \gain{4.3} &
  75.9 \gain{3.9} &
  90.8 \gain{4.4} &
  67.0 \gain{6.8} &
  75.6 \gain{4.0} \\ \bottomrule
\end{tabular}%
}
\end{table*}

\textbf{Efficacy of The Coarse-Grained Refinement} The results for the coarse refinement in Tab.~\ref{tab:corase filtering} show two clear trends: \textbf{a) the samples with old GUI patterns are harmful} and \textbf{b) REG task samples are also not helpful for enhancing GUI element grounding capability.}

Samples containing old GUI patterns are detrimental to performance. Their removal not only reduces fine-tuning costs but also leads to significant improvements across benchmarks.
This suggests a mismatch between outdated patterns and contemporary benchmarks, which are likely based on more recent digital interfaces.

Func. REG samples are the most harmful among the REG tasks.
Removing them brings notable gains on the benchmarks.
Adding Func, REG samples is harmful probably because it is hard for Florence-2 (the base model) to learn to generate detailed functional descriptions. Note that Florence-2 does not use a well-trained LLM as InternVL and Qwen2VL do.
Adding Text REG. and Brief Desc. REG samples are also not essential, probably because these REG tasks are dual to the original grounding tasks and share the same target elements, thereby providing no new training signals.

Overall, the results indicate the importance of selecting appropriate training samples, particularly avoiding outdated patterns and REG tasks, to enhance the model's capability of GUI element grounding.

\noindent \textbf{Efficacy of The Fine-Grained Refinement} Afterward, we investigate the efficacy of the fine refinement on the remaining grounding task samples.
As stated in Sec.~\ref{sec:core set curation}, this ablation study is conducted for the six main GUI metadata sources. Concretely, the samples from the main sources are merged to train GoClick as a baseline.
Afterward, the data inclusion ratio of each task type is decreased to observe the contribution of this type.
To easily discern overarching trends, the performance values on the grounding benchmarks are averaged.

The results in Fig.~\ref{fig:fine filtering} shows a clear trend: \textbf{the grounding task samples generated from the six main sources can lead to grounding accuracy gains but not all samples are useful, especially the samples generated from the redundant and overrepresented GUI metadata}.
Four sources, i.e. AutoGUI, MobileViews, MultiUI, and WebUI, show a clear positive correlation between inclusion ratio and accuracy for all task types.
Notably, for AutoGUI (Source 1), the convergence of all task types at ratio 1 indicates complementary benefits of these tasks.
MobileViews (Source 2) exhibits similar positive trends but with distinct differences between task types. Widget Listing tasks show a U-shaped curve, indicating potential noise at intermediate inclusion levels. In contrast, Text Grounding shows a consistent positive trend.
MultiUI (Source 5) shows divergent trends between task types. Brief Description Grounding maintains a slight positive trend, while Intent Grounding shows a dramatic improvement, suggesting that Intent Grounding samples from this source are particularly valuable.
WebUI (Source 6): All task types show positive trends, with Brief Description Grounding exhibiting the most substantial improvement, indicating that WebUI provides valuable Brief Description Grounding samples.

However, the two sources do not show consistently increasing trends:
AndroidControl (Source 3) exhibits a negative correlation for Text Grounding tasks.
This may be due to the less diverse GUI patterns presented in this source, compared with the more than 20,000 apps included in MobileViews (source 2).
Conversely, Functionality Grounding shows a positive trend, as these samples are annotated with LLM annotators~\citep{autogui}, thereby providing more meaningful supervision.
SeeClick-Web (Source 4) contains over 2 million Brief Description Grounding tasks but shows a catastrophic decline at full inclusion.
This precipitous drop suggests that including a large portion probably leads to overfitting as we find that this source contains too many pure-text webpages without pattern diversity.

These results empirically validate the importance of fine-grained data refinement in UI element grounding tasks and demonstrate that naive data aggregation can lead to deteriorated performance.

\noindent \textbf{Final Performance} Tab.~\ref{tab:data ablation overview} shows the efficacy of the two refinement stages. The coarse refinement leads to a notable performance gain and the fine-grained refinement brings a further improvement with significantly fewer training samples. The results highlight the importance of data refining when aggregating training data from diverse sources and validate the proposed core set curation pipeline.

\subsection{Experiment on GUI Agent Tasks via Device-Cloud Collaboration}
\label{sec:exp 2stage planning}

This subsection demonstrates a promising application of our lightweight grounding expert model, which is to collaborate with powerful proprietary models on cloud servers to solve GUI operation tasks on behalf of human users. The experimental settings, used benchmarks, compared methods, and experimental results are presented in Sec.~\ref{sec:agenttask exp setting}, Sec.~\ref{sec:gui agent bmks}, Sec.~\ref{sec:agenttask compared methods}, and Sec.~\ref{sec:2stage result}, respectively.

\subsubsection{Experimental Settings}
\label{sec:agenttask exp setting}
\begin{figure}[t]
    \centering
    \includegraphics[width=1\linewidth]{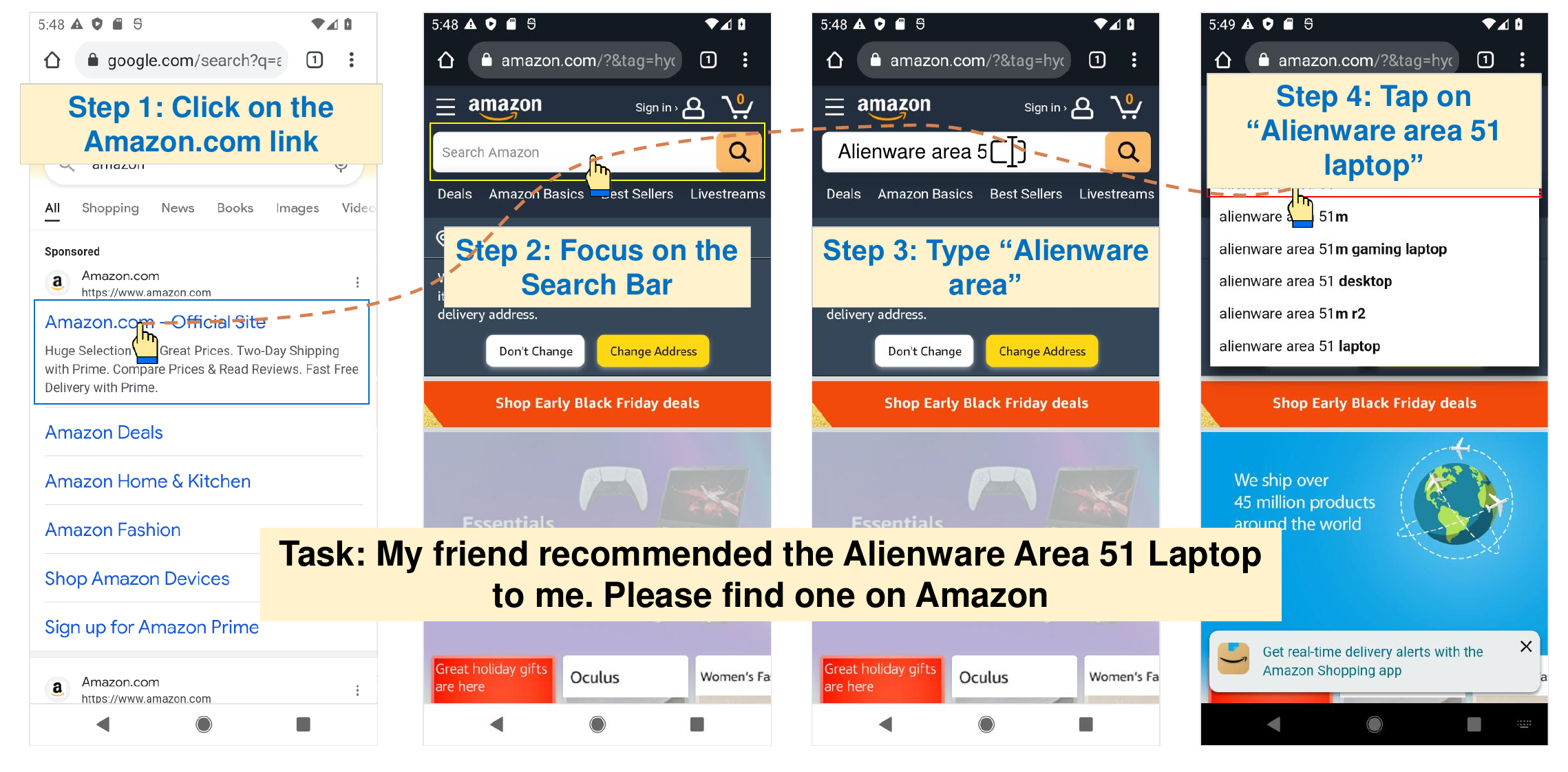}
    \caption{
    \textbf{Example of GUI agent tasks provided by the benchmarks in Sec.~\ref{sec:gui agent bmks}}. Each task contains a high-level user command, a sequence of recorded GUI screenshots, and associated ground truth actions. A GUI agent is required to output correct actions that match ground truths given the observations at every step.
    }
    \label{fig:gui agent task example}
\end{figure}

A GUI agent task requires an agent to operate GUIs, such as mobile apps, to complete high-level instructions issued by human users (examples in Fig.~\ref{fig:gui agent task example}). As the GUI metadata of apps are not always available due to anti-crawler mechanisms deployed on contemporary apps, the environmental observation of the agent in this experiment only includes GUI screenshot images without accessing the node attributes recorded in GUI source code. This means that the agent model should plan actions based on solely visual observations.

\begin{figure*}[t]
    \centering
    \includegraphics[width=1\linewidth]{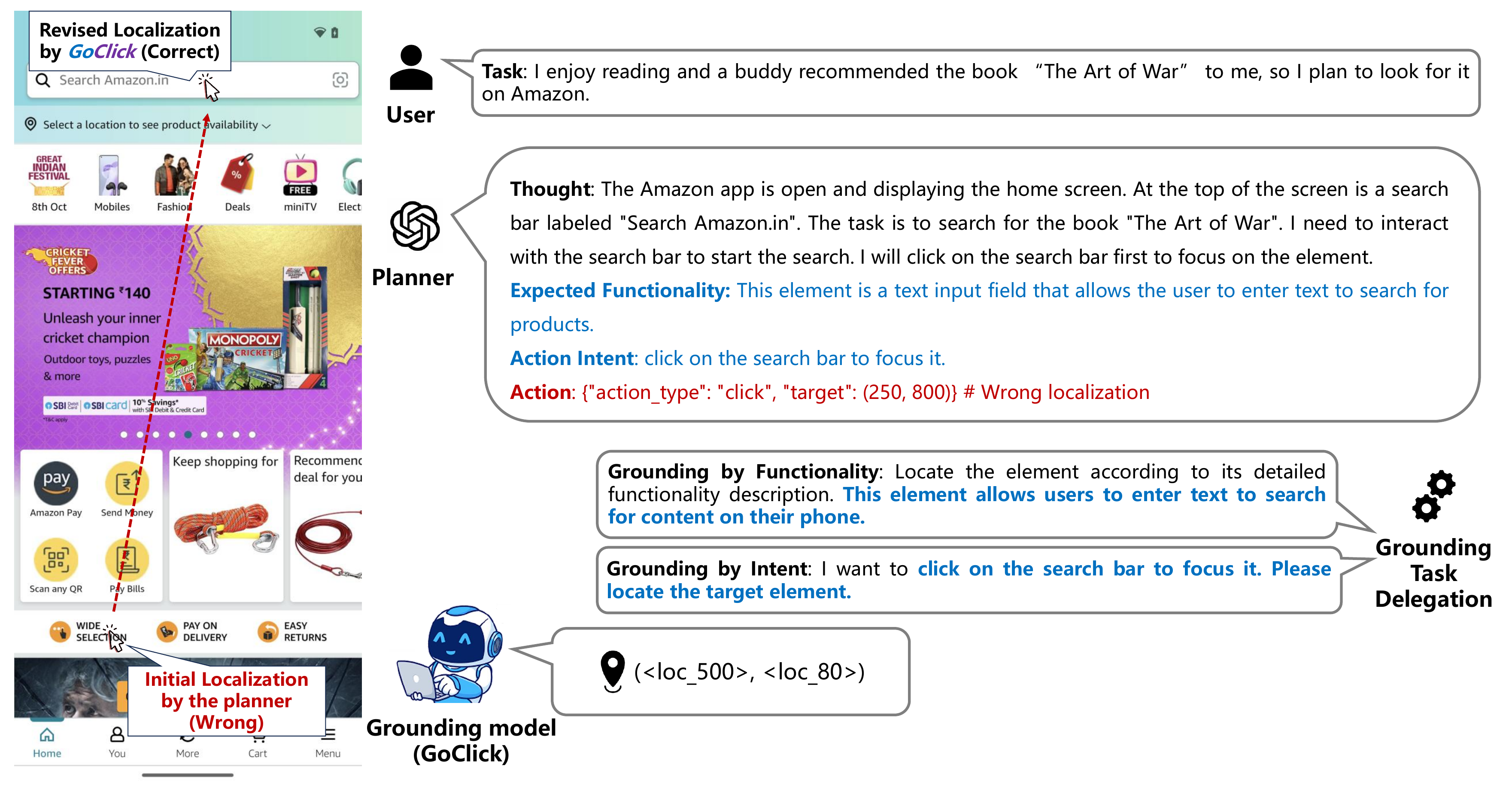}
    \caption{
    \textbf{Illustration of the device-cloud collaboration agent integrating our GoClick}. We adopt a 2-stage planning pipeline, that prompts cloud-based proprietary models (e.g., GPT-4o) to conduct reasoning and planning (planner) and then delegates element grounding to GoClick (grounding model). While proprietary models excel at reasoning and task planning, they may not be good at outputting coordinates of the targets in GUI element grounding tasks, as shown in the experiments in Sec.~\ref{sec:2stage result}. GoClick can overtake element grounding and helps the agent improves task success rate, demonstrating the potential application of GoClick.
    }
    \label{fig:2stage example}
\end{figure*}

In this experiment, the agent integrates one \textbf{\textit{task planner}} deployed on cloud servers and one \textbf{\textit{visual grounding model}} deployed on personal mobile phones. The task planner based on proprietary VLMs (e.g., GPT-4o) on cloud servers plan the next action by either outputting action intents or describing the expected functionality requirement that the target element should fulfill.
Subsequently, the grounding model (i.e., our GoClick) takes the same GUI screenshot image as input to locate the target element according to the action intent or functionality description in textual format. An example is illustrated in Fig.~\ref{fig:2stage example}.

This 2-stage agent task setting is adopted because: (1) Proprietary VLMs, such as GPT-4o, probably possess weak GUI element grounding capabilities (shown in the experimental results) despite general reasoning capability; (2) Our GoClick is a grounding expert model without free-form reasoning ability and needs to collaborate with a planner that output accurate referring expressions of task-related target elements.

In real deployment, the on-device process can use an HTTP request library to send the user task and GUI screenshots to the planner servers and fetch the planner’s inference results. The atomic action extracted from the inference result is then implemented by the Android Debugging Bridge to emulate human behavior, such as click, swipe, and type. However, due to the long time used for online evaluation on real edge devices, we adopt the offline evaluation protocol typically employed in recent GUI agent works, such as AndroidControl~\citep{androidcontrol} and Mind2Web~\citep{deng2024mind2web}. This offline evaluation stores the testing trajectories and assesses the predictions of the tested agents given the observation inputs of all steps. To accelerate evaluation, asynchronous processing is adopted to handle batch inference. The evaluation details will be introduced for each used benchmark in Sec.~\ref{sec:gui agent bmks}.

\subsubsection{GUI Agent Task Benchmarks}
\label{sec:gui agent bmks}
This subsection introduces the pixel-based GUI agent task benchmarks and their evaluation metrics.

\begin{enumerate}
    \item \textbf{Android-in-The-Wild (AITW)}~\citep{rawles2023android}: This is a large Android phone control dataset covering multiple scenarios such as app Store operation, browser use, and web shopping, on more than 350 Apps. As this dataset provides redundant ground truth trajectories for each task, testing on all trajectories is infeasible due to the time limit. Therefore, we use the test split (4663 samples unfolded from 584 trajectories) provided in SeeClick~\citep{cheng2024seeclick}, which retains only one trajectory for each task to guarantee efficient evaluation.
    \item \textbf{AndroidControl}~\citep{androidcontrol}: This dataset contains 15K unique tasks over 833 Apps. Its test split provides 12685 samples unfolded from 1479 trajectories. The ground truth actions are annotated by human experts for evaluation. We follow the setting in AndroidControl~\citep{androidcontrol} to use a random-500-sample split for testing.
    \item \textbf{GUIAct}~\citep{guicourse}: This benchmark provides more difficult tasks, such as information-seeking tasks with complex user preferences. The tasks on mobile and web domains are both provided by this benchmark.

\end{enumerate}

% \usepackage{booktabs}
% \usepackage{graphicx}
\begin{table*}[]
\centering
\caption{
\textbf{Evaluating the device-cloud collaboration agent integrating GoClick on the AITW benchmark}~\citep{rawles2023android}. Apart from the major \textbf{Step SR} metric, \textbf{Accuracy of Click Action} is also reported in parentheses. Prompting the proprietary models to conduct both planning and grounding leads to inferior Click Accuracy and Step SR (row 1). Using SoM prompting can enhance the performance (row 2). Delegating the element grounding to GoClick by providing either the action intent or expected functionality output by the planner significantly increases Click Accuracy and Step SR.}
\label{tab:2stage aitw}
\resizebox{\textwidth}{!}{%
\begin{tabular}{@{}cc|cccccc@{}}
\toprule
Planner                  & Grounding Model         & General      & Install     & Google Apps & Single      & Web shopping & Overall     \\ \midrule

Gemini-2-Flash-Exp       & -                       & 26.4 (18.2)  & 28.5 (26.9) & 30.3 (22.9) & 41.9 (29.0) & 20.2 (22.7)  & 29.5 (23.6) \\

Gemini-2-Flash-Exp + SoM & -                       & 29.9 (32.2)  & 33.9 (41.4)     & 33.9 (30.1)   & 48.5 (56.8)   &  27.9 (37.8)   &  34.8 (38.3)   \\
Gemini-2-Flash-Exp       & GoClick-L w/ Intent Gnd. (ours) & 43.1 (48.1)  & \textbf{44.1} (52.9) & \textbf{49.5} (54.6) & \textbf{59.7} (67.4) & 39.8 (54.8)  & \textbf{47.2} (54.0) \\
Gemini-2-Flash-Exp       & GoClick-L w/ Func. Gnd. (ours) & \textbf{43.2} (48.4)  & 40.9 (47.4) & 48.5 (52.9) & 59.4 (66.8) & \textbf{40.0} (55.0)  & 46.4 (52.5) \\ \midrule

GPT-4o                   & -                       & 28.2 (24.9)  & 32.9 (30.1) & 31.9 (27.6) & 44.2 (47.1) & 25.9 (30.0)  & 27.2 (29.9) \\

GPT-4o + SoM             & -                       & 33.7 (37.2)  & 43.2 (53.8) & 41.4 (51.8) & 53.0 (63.2) & 39.2 (51.0)  & 42.1 (50.4) \\
GPT-4o                   & GoClick-L w/ Intent Gnd. (ours) & \textbf{45.9} (57.4)  & \textbf{50.0} (59.0) & \textbf{49.7} (57.1) & \textbf{54.4} (69.7) & \textbf{44.5} (60.5)  & \textbf{48.9} (59.7) \\
GPT-4o                   & GoClick-L w/ Func. Gnd.  (ours) & 45.7 (56.9)  & 47.1 (51.2) & 47.9 (54.1) & 53.5 (67.4) & 44.0 (59.7)  & 47.6 (57.5) \\ \bottomrule
\end{tabular}%
}
\end{table*}
% \usepackage{booktabs}
% \usepackage{graphicx}
\begin{table}[]
\centering
\caption{
\textbf{Evaluating the device-cloud collaboration agent integrating GoClick on the AndroidControl benchmark}~\citep{androidcontrol}.
Similar to the trend on AITW, delegating the element grounding to GoClick also significantly increases Click Accuracy and Step SR on this benchmark, outperforming the proprietary models equipped with SoM prompting.}

\label{tab:2stage ancon}
\resizebox{\linewidth}{!}{%
\begin{tabular}{@{}cc|cc@{}}
\toprule
Planner                  & Grounding Model         & Step SR $\uparrow$ & Click Acc. $\uparrow$\\ \midrule
Gemini-2-Flash-Exp       & -                       & 20.6    & 11.4       \\

Gemini-2-Flash-Exp + SoM & -                       & 35.3    & 44.8       \\
Gemini-2-Flash-Exp       & GoClick-L w/ Intent Gnd. (ours) & \textbf{42.9}    & \textbf{49.8}       \\
Gemini-2-Flash-Exp       & GoClick-L w/ Func. Gnd. (ours) & 41.9    & 48.3       \\ \midrule

GPT-4o                   & -                       & 19.5    & 14.0       \\

GPT-4o + SoM             & -                       & 39.0    & 48.3       \\
GPT-4o                   & GoClick-L w/ Intent Gnd. (ours) & \textbf{42.5}    & \textbf{53.3}       \\
GPT-4o                   & GoClick-L w/ Func. Gnd. (ours) & 41.9    & 52.2       \\ \bottomrule
\end{tabular}%
}
\end{table}
% \usepackage{booktabs}
% \usepackage{graphicx}
\begin{table}[]
\centering
\caption{\textbf{Evaluating the device-cloud collaboration agent integrating GoClick on the GUIAct-Mobile benchmark}~\citep{guicourse}. On this challenging benchmark, the proprietary VLMs perform poorly. Delegating the element grounding task to our GoClick substantially improves the Step SR and Click Accuracy, outperforming the SoM prompting method.}
\label{tab:2stage guiactmobile}
\resizebox{\columnwidth}{!}{%
\begin{tabular}{@{}cc|cc@{}}
\toprule
Planner                  & Grounding Model                 & Step SR $\uparrow$       & Click Acc. $\uparrow$    \\ \midrule
Gemini-2-Flash-Exp       & -                               & 19.6          & 17.6          \\
Gemini-2-Flash-Exp + SoM & -                               & 23.3          & 25.2          \\
Gemini-2-Flash-Exp       & GoClick-L w/ Intent Gnd. (ours) & \textbf{28.7} & \textbf{28.6} \\
Gemini-2-Flash-Exp       & GoClick-L w/ Func. Gnd. (ours) & 27.2          & 26.1          \\ \midrule
GPT-4o                   & -                               & 28.1          & 28.8          \\
GPT-4o + SoM             & -                               & 27.2          & 28.6          \\
GPT-4o                   & GoClick-L w/ Intent Gnd. (ours) & \textbf{34.6} & \textbf{29.6} \\
GPT-4o                   & GoClick-L w/ Func. Gnd. (ours)  & 34.2          & 28.8          \\ \bottomrule
\end{tabular}%
}
\end{table}

These benchmarks assess agents' ability to predict the next actions given user tasks, GUI screenshot images, and textual action history, without being actually deployed on either physical devices or emulators.
Following the popular evaluation setting~\citep{osatlas,androidcontrol, aguvis}, past GUI screenshot images are not included in the prompt for the planning models due to a context length limit.

The major metric is \textbf{Step Success Rate (Step SR)}~\citep{deng2024mind2web,rawles2023android}, defined as the proportion of steps for which all action parameters match the ground truth.
\begin{equation}
\text{Step SR} = \frac{\sum_{i=1}^{N} \delta(\text{Action}_i, \text{GT}_i)}{N} \times 100
\end{equation}

where \( N \) is the total number of steps, \(\text{Action}_i\) represents the action parameters for the \(i\)-th step, \(\text{GT}_i\) represents the ground truth parameters for the \(i\)-th step, and \(\delta(\text{Action}_i, \text{GT}_i)\) is an indicator function defined as:

\begin{equation}
\delta(\text{Action}_i, \text{GT}_i) =
\begin{cases} 
1 & \text{if } \text{Action}_i = \text{GT}_i \\
0 & \text{otherwise}
\end{cases}
\end{equation}

For example, an action \textit{click(target\_x, target\_y)} is correct only if the action type matches the ground truth and the predicted target coordinates fall within the ground truth bounding box.
An action \textit{input\_text(text)} is correct only if the F1 score of the predicted text parameter is greater than a benchmark-specific threshold (e.g., 0.5 in AndroidControl~\citep{androidcontrol}).

As the samples that require element grounding (e.g. click and long-press) occupy 57.8\% of AITW, 55.6\% of AndroidControl, and 61.2\% of GUIAct, performance gains brought by GoClick overtaking element grounding will be notably reflected on the final Step SR.

\subsubsection{Compared Methods}
\label{sec:agenttask compared methods}
\noindent \textbf{Simple Baseline: Proprietary VLMs that conduct both planning and grounding } This method prompts a proprietary VLM to generate action plans and then to predict the coordinates of the target element required by click/long-press actions.

Besides predicting actions, the proprietary VLMs also output the expected functionality descriptions of the target elements and summarize action intents.
Our device-cloud collaboration agent will delegate the element grounding task to GoClick by prompting GoClick to locate target elements according to the functionality descriptions and action intents.

\noindent \textbf{Strong Baseline: Proprietary VLMs using Set-of-Marks Prompting (SoM)} This method employs an element detection model by OmniParser~\citep{omniparser} to detect all elements on a GUI screenshot image, labels them with bounding boxes and numeric tags in different colors following SoM~\citep{som} and prompts a proprietary model to choose the target element's tag from the candidates marked on the image. This method is a strong baseline as it reduces the huge output space of coordinates to dozens of element candidates and casts the difficult grounding task into a multi-choice problem.

\subsubsection{Experimental Results}
\label{sec:2stage result}

% \usepackage{booktabs}
% \usepackage{graphicx}
\begin{table}[]
\centering
\caption{\textbf{Evaluating the device-cloud collaboration agent integrating GoClick on the GUIAct-Web benchmark}~\citep{guicourse}. Apart from the major \textbf{Step SR} metric, \textbf{Accuracy of Click Action} is also reported in parentheses. Although using Set-of-Marks prompting~\citep{som} can enhance the performance of the proprietary VLMs, this is still inferior to the approach that delegates the element grounding to our GoClick.}
\label{tab:2stage guiactweb}
\resizebox{\columnwidth}{!}{%
\begin{tabular}{@{}cc|cc@{}}
\toprule
Planner                  & Grounding Model                 & Step SR $\uparrow$      & Click Acc. $\uparrow$   \\ \midrule
Gemini-2-Flash-Exp       & -                               & 16.8          & 8.0           \\
Gemini-2-Flash-Exp + SoM & -                               & 32.9          & 44.7          \\
Gemini-2-Flash-Exp       & GoClick-L w/ Intent Gnd. (ours) & \textbf{41.7} & \textbf{51.6} \\
Gemini-2-Flash-Exp       & GoClick-L w/ Func. Gnd. (ours) & 39.9          & 48.5          \\ \midrule
GPT-4o                   & -                               & 18.2          & 5.1           \\
GPT-4o + SoM             & -                               & 42.3          & 55.6          \\
GPT-4o                   & GoClick-L w/ Intent Gnd. (ours) & \textbf{50.5} & \textbf{62.0} \\
GPT-4o                   & GoClick-L w/ Func. Gnd. (ours)  & 47.8          & 57.2          \\ \bottomrule
\end{tabular}%
}
\end{table}

The experimental results in Tab.~\ref{tab:2stage aitw}, Tab.~\ref{tab:2stage ancon}, Tab.~\ref{tab:2stage guiactmobile}, and Tab.~\ref{tab:2stage guiactweb} demonstrate the effectiveness of the two-stage planning approach that combines powerful proprietary models for planning with the specialized GoClick model for visual element grounding. Several key observations can be made:

\noindent \textbf{A) Performance Improvement with GoClick.} When comparing the simple baseline against the approach integrating with GoClick, we observe substantial performance gains across all benchmarks:

On AITW, for Gemini-2-Flash-Exp, the overall Step Success Rate (Step SR) improves from \textbf{29.5} to \textbf{47.2} when combined with GoClick using Intent Grounding, showing an improvement of \textbf{17.7}. This improvement is attributed to the significantly higher element grounding accuracy, with an increase by \textbf{30.4}, from \textbf{23.6} to \textbf{54.0}.

For GPT-4o, the overall Step SR improves from \textbf{27.2} to \textbf{48.9} when combined with GoClick using Intent Grounding. This improvement is also attributed to the nearly doubled element grounding accuracy (\textbf{29.9} $\rightarrow$ \textbf{59.7}).

On AndroidControl, significant performance gains are also witnessed, with a \textbf{22.3 (20.6 $\rightarrow$ 42.9}) and \textbf{23.0 (19.5 $\rightarrow$ 42.5)} increase for Gemini-2-Flash-Exp and GPT-4o, respectively.

The same trends can also be seen on the more challenging GUIAct-Web and GUIAct-Mobile, demonstrating that equipping proprietary VLM planners with GoClick leads to higher Step SR.

\noindent \textbf{B) Integrating with GoClick, with either intent grounding or functionality grounding, outperforms the competitive SoM prompting strategy.}
On AITW, Gemini-2-Flash-Exp + SoM achieves 34.8 overall Step SR, while the agent integrating GoClick reaches a higher \textbf{47.2}. GPT-4o + GoClick also outperforms GPT-4o + SoM.

On AndroidControl and the two GUIAct benchmarks, integrating with GoClick also surpasses the strong SoM baseline.

We note that the SoM baseline also needs an expert model to help proprietary VLMs better perform element grounding in GUI agent tasks, but the results show that this is inferior to delegating element grounding to our lightweight GoClick.

\noindent \textbf{C) Intent Grounding vs. Functionality Grounding.}
Element grounding tasks can be delegated to GoClick either by Intent Grounding or Functionality Grounding. The results show slight differences between the two approaches.
For example, on AITW, Intent Grounding slightly outperforms Functionality Grounding. A similar trend is also seen for GPT-4o. On the other three benchmarks, this slight gap also exists.

These differences suggest that using GoClick to conduct action intent grounding may be marginally more effective than grounding by functionality descriptions. This is likely because locating an element given its indirect functional description is harder than locating it given its direct appearance and positional description.

In summary, the device-cloud collaboration combining proprietary VLMs for planning with the specialized GoClick model for visual element grounding significantly outperforms standalone proprietary VLMs across the GUI agent task benchmarks.
This suggests that addressing the visual grounding bottleneck through our specialized model while leveraging the reasoning capabilities of proprietary models creates a synergistic effect, establishing a promising direction for developing more effective GUI agents that can accurately interpret and interact with GUIs on behalf of users.
%%%% Section: Conclusion
\section{Conclusion And Limitations}\label{sec:conclusion}
This paper introduces GoClick, a lightweight VLM that achieves state-of-the-art GUI element grounding accuracy while remaining small enough for on-device deployment.
First, we build GoClick based on an encoder-decoder architecture that outperforms decoder-only alternatives at small parameter scales for GUI grounding tasks.
Second, we propose and validate a systematic data curation pipeline that combines coarse refinement and fine-grained task ratio adjustment. This approach not only reduces training costs but significantly improves grounding accuracy by removing harmful and redundant samples.
Third, we show that integrating lightweight on-device grounding models like GoClick into device-cloud collaboration frameworks enhances overall GUI agent performance, offering a practical pathway to deploying intelligent GUI assistants on resource-constrained devices.

Despite these contributions, our work still possesses several limitations:
a) GoClick's architecture choice is optimized for GUI element grounding, and the encoder-decoder advantage we observe may not generalize to other GUI-related tasks. Future work can explore whether these architectural insights transfer to more complex tasks, such as GUI agent task planning, action prediction, and even challenging Chain-of-Thoughs~\citep{wei2022chain}, under similar parameter constraints.

b) While our progressive data refinement pipeline yields significant improvements, it remains partially heuristic. In addition, repeated experiments are also hard to conduct due to computation resource limits, which may lead to slight variations in the data ratio adjustment experiments when using different random seeds to extract subsets. A more theoretically grounded approach to identifying harmful training samples could further enhance performance.
Additionally, our finding that outdated GUI patterns degrade performance indicates that GoClick may need periodic retraining as GUI design evolves.

We believe these future directions can unleash the full power of our PDR method: (a) Meta-Learning-Based: Following recent work on data quality prediction~\citep{xu2024demystifying}, we could train a meta-model that predicts sample quality based on features such as GUI age, element density, RE complexity, and source distribution. This model could be trained using our ablation results as a supervision signal. (b) Game Theory-Based: Techniques like Shapley values or influence functions could automatically identify high-value samples by measuring their contribution to validation performance. However, this approach is computationally expensive—computing influence for 10.8M samples on our 0.8B model would require substantial resources.

c) Our evaluation is performed on L20 GPUs rather than real embedded devices. As no mature methods of deploying VLMs on embedded systems are available, we simulate resource constraints in a controlled environment where performance can be reliably measured and compared. While necessary for systematic evaluation, this setting may not fully capture the challenges of real-world embedded deployment, including memory constraints, energy consumption, and system-specific optimizations. Future work can focus on developing specialized deployment frameworks for VLMs on embedded systems and validating GoClick's performance under these authentic conditions.

We hope these findings offer implications for the deployment of multimodal GUI agents in resource-constrained environments.

\section*{Declarations}

\noindent\textbf{Acknowledgements}
This work was supported in part by the National Key R\&D Program of China (No. 2022ZD0160102), the National Natural Science Foundation of China (No. U21B \\
2042, No. 62320106010), and in part by the 2035 Innovation Program of Chinese Academy of Sciences.

\noindent\textbf{Data Availability.}
\justifying
The datasets used to conduct data refinement experiments and to train GoClick can be accessed at \url{https://github.com/ZJULiHongxin/GoClick}.

The GUI agent task testing sets are also available at \url{https://github.com/ZJULiHongxin/GoClick}.

The GUI element grounding benchmarks used in our experiments are available from FuncPred\footnote{https://huggingface.co/datasets/AutoGUI/AutoGUI-v1-test}, ScreenSpot\footnote{https://huggingface.co/datasets/rootsautomation/ScreenSpot}, Screen-Spot-v2\footnote{https://huggingface.co/datasets/HongxinLi/ScreenSpot\_v2}, MOTIF\footnote{https://huggingface.co/datasets/HongxinLi/MOTIF-EVAL}, RefExp\footnote{https://huggingface.co/datasets/ivelin/ui\_refexp\_saved}, VWB AG\footnote{https://huggingface.co/datasets/HongxinLi/VWB-AG}, and VWB EG\footnote{https://huggingface.co/datasets/HongxinLi/VWB-EG}.

{\footnotesize
    \bibliographystyle{apalike}
    \bibliography{ref.bib}
}

\end{document}